\documentclass{article}
\usepackage{iclr2027_conference,times}
\usepackage[scaled=.94]{helvet}

\usepackage{amsmath,amsfonts,bm}

\def\eqref#1{equation~\ref{#1}}

\def\1{\bm{1}}

\DeclareMathAlphabet{\mathsfit}{\encodingdefault}{\sfdefault}{m}{sl}
\SetMathAlphabet{\mathsfit}{bold}{\encodingdefault}{\sfdefault}{bx}{n}

\usepackage{amsmath,amssymb,mathtools}
\usepackage{booktabs,multirow}
\usepackage{graphicx}
\graphicspath{{./}}
\usepackage{placeins}
\usepackage{float}
\usepackage{xcolor}
\usepackage{microtype}
\usepackage{tikz}
\usetikzlibrary{arrows.meta,positioning,fit,backgrounds,calc,decorations.pathmorphing}
\usepackage{hyperref}
\usepackage{url}
\usepackage{etoolbox}
\makeatletter
\patchcmd{\@maketitle}{\vskip 0.3in minus 0.1in}{\vskip 8pt}{}%
  {\PackageError{arxiv-layout}{Title spacing patch failed}{}}
\patchcmd{\@maketitle}{\rule{\z@}{24pt}}{\rule{\z@}{16pt}}{}%
  {\PackageError{arxiv-layout}{Author spacing patch failed}{}}
\makeatother
\renewenvironment{abstract}{%
  \par\vspace{5pt}\centerline{\large\scshape Abstract}\vspace{0.5ex}%
  \begin{list}{}{\setlength{\leftmargin}{0pt}%
    \setlength{\rightmargin}{0pt}\setlength{\topsep}{4pt}}%
  \item\relax
}{\end{list}\vspace{4pt}}

\definecolor{slowred}{HTML}{D66555}
\definecolor{slowredlight}{HTML}{F3D6D1}
\definecolor{fastblue}{HTML}{4F75B3}
\definecolor{fastbluelight}{HTML}{D8E1F0}
\definecolor{ambigpurple}{HTML}{9575B5}
\definecolor{ambigpurplelight}{HTML}{E5DCEF}
\definecolor{causalteal}{HTML}{3A9688}
\definecolor{causalteallight}{HTML}{D8ECE8}
\definecolor{commitamber}{HTML}{D19A45}
\definecolor{commitamberlight}{HTML}{F2E4C9}

\definecolor{primaryink}{HTML}{30343B}
\definecolor{secondarytext}{HTML}{899097}
\definecolor{darksecondary}{HTML}{626970}
\definecolor{structurefill}{HTML}{F1F0ED}
\definecolor{targetfill}{HTML}{E3E7E8}
\definecolor{warmcard}{HTML}{F8F6F1}
\definecolor{inactivegray}{HTML}{C8CBCB}
\definecolor{gridline}{HTML}{DDE0E2}
\definecolor{cardborder}{HTML}{D9D7D2}
\definecolor{paperwhite}{HTML}{FFFFFF}

\definecolor{huezero}{HTML}{D66555}
\definecolor{hueone}{HTML}{C96868}
\definecolor{huetwo}{HTML}{BC6B7B}
\definecolor{huethree}{HTML}{AF6F8F}
\definecolor{huefour}{HTML}{A272A2}
\definecolor{huefive}{HTML}{9575B5}
\definecolor{huesix}{HTML}{8775B5}
\definecolor{hueseven}{HTML}{7975B4}
\definecolor{hueeight}{HTML}{6B75B4}
\definecolor{huenine}{HTML}{5D75B3}
\definecolor{hueten}{HTML}{4F75B3}

\tikzset{
  paperpanel/.style={draw=cardborder, rounded corners=4pt, fill=paperwhite,
    line width=.6pt},
  naturalcard/.style={draw=cardborder, rounded corners=3pt, fill=warmcard,
    line width=.8pt, inner xsep=5pt, inner ysep=4pt, align=center},
  conditiontoken/.style={draw=cardborder, rounded corners=2pt, fill=structurefill,
    line width=.8pt, inner xsep=4pt, inner ysep=3pt, align=center},
  targettoken/.style={draw=cardborder, rounded corners=2pt, fill=targetfill,
    line width=.8pt, inner xsep=4pt, inner ysep=3pt, align=center},
  causalcard/.style={draw=causalteal, rounded corners=3pt, fill=causalteallight,
    line width=1pt, inner xsep=5pt, inner ysep=4pt, align=center},
  commitcard/.style={draw=commitamber, rounded corners=3pt, fill=commitamberlight,
    line width=1pt, inner xsep=5pt, inner ysep=4pt, align=center},
  naturalarrow/.style={-{Latex[length=2mm]}, draw=primaryink, line width=.9pt},
  causalarrow/.style={-{Latex[length=2mm]}, draw=causalteal, line width=1.25pt},
  association/.style={draw=secondarytext, densely dotted, line width=.8pt},
  boundaryline/.style={draw=commitamber, line width=1.25pt}
}

\newcommand{\resultlead}[1]{\par\noindent{\bfseries #1}\hspace{0.35em}}
\newcommand{\introhighlight}[1]{{\bfseries #1}}

\title{A Chosen Future Can Still Be Rewritten:\\
Causal Writability in Video Models}

\author{%
\parbox[t]{\dimexpr\textwidth-2\tabcolsep\relax}{\centering
Xingyun Wang\textsuperscript{1,4,*}\hfill
Haomin Zheng\textsuperscript{2,4,*}\hfill
Man Yuan\textsuperscript{2,4}\hfill
Leqian Yang\textsuperscript{3,4}\hfill
Ziming Liu\textsuperscript{1,4,5}\\[4pt]
{\normalfont\small
\textsuperscript{1}Tsinghua University\quad
\textsuperscript{2}Peking University\quad
\textsuperscript{3}University of Science and Technology of China\\
\textsuperscript{4}MetaCircle\qquad
\textsuperscript{5}Shanghai Qi Zhi Institute\\[2pt]
\textsuperscript{*}Equal contribution.\quad
\href{mailto:xingyun-24@mails.tsinghua.edu.cn}{xingyun-24@mails.tsinghua.edu.cn}\quad
\href{mailto:zmliu@tsinghua.edu.cn}{zmliu@tsinghua.edu.cn}\\[2pt]
\href{https://xingyun-24.github.io/causal-writability/}{Project page}\qquad
\href{https://github.com/xingyun-24/causal-writability}{GitHub}
}}
}

\iclrfinalcopy
\begin{document}
\raggedbottom
\maketitle
\lhead{Preprint}

\begin{abstract}
When a video model generates physically incorrect motion, did it fail to learn
the correct motion, or did it learn it but fail to use it?
\textbf{We show the latter: the correct motion remains available inside the
model and can still be made to control the generated video.} We train on videos
where red masses oscillate slowly and blue masses oscillate quickly, then test a
red mass with fast observed motion. Even when the model generates slow motion in
this conflicting case, a low-dimensional edit predicted from simple physical
variables restores the correct fast motion. \textbf{We call this ability causal
writability. At fixed strength, we find a sharp depth boundary: the same edit
changes the video before the boundary but not after it.} This closure marks
commitment for that write. The motion signal nevertheless remains, and a
stronger downstream write can restore physical motion, while excessive gain
overshoots. \textbf{Early causal writability predicts which errors training
later corrects:} those errors are writable at more network depths than errors
that persist. We reproduce both causal writability and its sharp closure in a
pretrained 1.3B video model, supporting generality across model scale and
training regime.
\end{abstract}

\begingroup
\setlength{\intextsep}{8pt}
\begin{figure}[H]
    \centering
    \includegraphics[trim=0 9.32pt 0 0,clip,width=\textwidth]{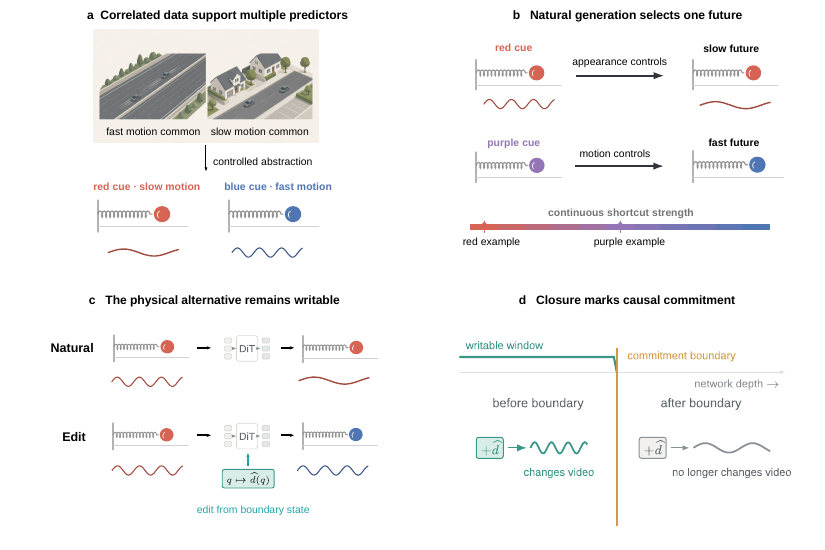}
    \caption{\textbf{Natural generation can select one future while a causally
    usable physical alternative remains writable.}
    \textbf{a,} Correlated red--slow and blue--fast examples support both
    motion-based and appearance-based predictive rules.
    \textbf{b,} With observed motion fixed, cue strength changes which future
    controls natural generation.
    \textbf{c,} An activation edit predicted from target motion and boundary
    state gives the physical alternative control over decoded video.
    \textbf{d,} Closure marks the depth after which the same edit no longer
    changes the decoded video.}
    \label{fig:visual-thesis}
\end{figure}
\endgroup
\clearpage

\section{Introduction}

\introhighlight{When a video model generates physically incorrect motion, has it failed
to learn the correct motion, or has it learned it but failed to use it?} The
generated video alone cannot distinguish these two failures. \introhighlight{We show
that the correct motion can remain available inside the model even when it does
not control the natural rollout; an intervention can give it control.}

Correlations in observational video create this ambiguity.
Generative world models learn from video to predict how scenes will evolve
\citep{ha2018worldmodels,hafner2020dream,ho2022video}, but appearance, context,
object identity, and motion rarely vary independently. Highway scenes, for
example, tend to contain faster vehicle motion, while residential scenes tend to
contain slower motion. Scene appearance can therefore predict speed statistically
without determining it physically. A model may extrapolate the future from recent
motion or use appearance as a proxy for speed. Ordinary training examples do not
distinguish these rules; only when appearance and observed motion conflict do they
predict different futures. This is predictive underspecification
\citep{geirhos2020shortcut,damour2022underspecification}.

To test whether the correct motion is absent or merely unused, we construct a
controlled spring--mass task. During training, red masses oscillate slowly and
blue masses oscillate quickly. At test time, we create unseen color--motion
combinations by holding observed motion fixed and changing only color---for
example, a red mass with fast observed motion. We determine which cue controls
the generated future by measuring motion directly in fully decoded videos.
\introhighlight{The model can follow color and generate slow motion---a predictive
shortcut.} A low-dimensional internal edit predicted from simple physical
variables restores the correct fast motion. Figure~\ref{fig:visual-thesis}
summarizes the conflict and the causal tests that follow.

\introhighlight{Causal writability is the ability of an internal edit to change motion
in decoded video.} It depends on both edit location and strength. As computation
proceeds, observed-frame information is progressively written into future-frame
states, so observed-frame edits can eventually arrive too late. At fixed
strength, we find a sharp depth boundary over a narrow range of layers: the same
edit changes the video before the boundary but not after it. Closure marks
commitment for that write. The underlying motion signal remains detectable after
commitment; an appropriate downstream gain restores physical motion, while
excessive gain overshoots.

\introhighlight{Early causal writability predicts which errors training later corrects.}
At an early checkpoint, those errors are writable at more network depths than
errors that persist. The writable boundary also depends on evidence and learning
history: more observed motion keeps the correct future writable to deeper layers,
and, in a pretrained 1.3B video model, learning motion before introducing the
color bias also moves the boundary deeper.

\introhighlight{Different natural behavior can coexist with a shared causal edit space.}
Independently trained models differ in how often their rollouts follow observed
motion, yet corrective edits transfer between them. This points to different
downstream use of similar physical solutions. We therefore trace the downstream
write and show that self-attention carries motion information from observed
frames into future-frame states; strengthening the corresponding write restores
the correct motion in decoded video.

\section{How Appearance and Observed Motion Compete for the Generated Future}
\label{sec:solution-selection}

\resultlead{Experimental setting.}
We study Spring, a 488M-parameter latent flow-matching Transformer with 30
bidirectional DiT blocks and a frozen Wan2.1 VAE. Each 20-fps video shows a
colored mass oscillating on a spring, with 65 observed frames followed by 64
generated frames. Training pairs red with the slow band
$\omega\sim U[2.2,3.0]$ rad/s and blue with the fast band
$\omega\sim U[5.2,6.4]$ rad/s. No conflicting color--frequency combinations
appear during training. We train two settings with different amounts of visible
motion. Long retains all 65 observed frames, whereas Short replaces the first 57
frames with a fixed background, leaving only the final 8 motion frames while
preserving the timeline and token count. We detect the mass in fully decoded
videos and measure its generated frequency and future appearance. Full model,
sampling, and evaluation details appear in
Appendix~\ref{app:experimental-details}.

\resultlead{Appearance-cue strength shifts the generated future.} We fix observed
motion and generation seed for each trajectory and sweep 11 hues from red to blue.
\textbf{Physics-follow rate} is the fraction of valid decoded videos whose
frequency lies in the observed-motion band, evaluated over the stated cohort.
Aggregate behavior pools the $64\times11$ grid (32 slow and 32 fast histories);
cue-specific and conflict-only rates use subsets.
The training color--frequency rule gives a 50\% reference on the full balanced
grid, but 0\% on conflict-only endpoints.

\begin{figure*}[t]
  \centering
  \includegraphics[width=.915\textwidth]{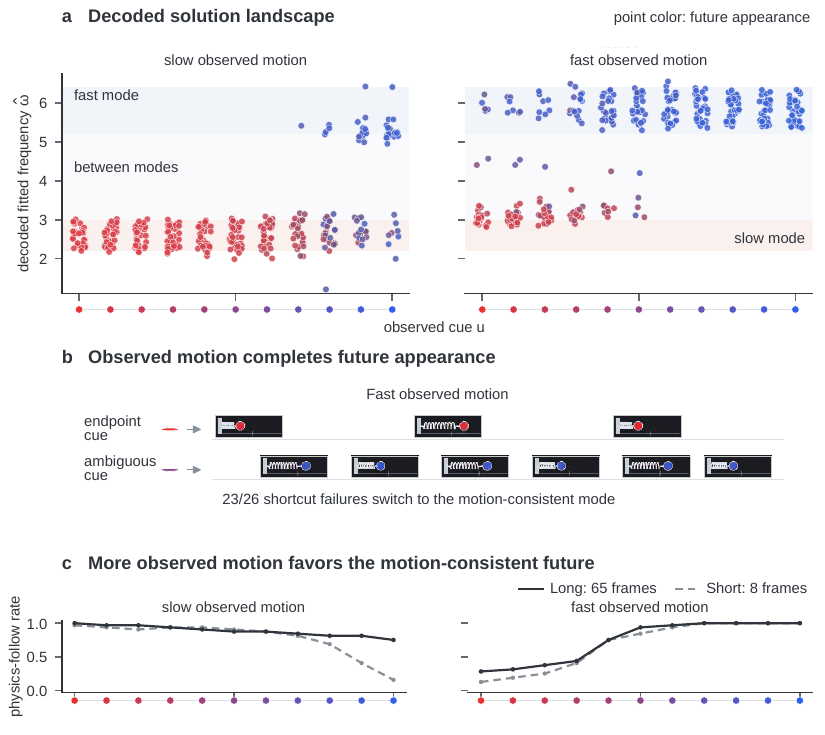}
  \caption{\textbf{Appearance and observed motion compete to determine the
  generated future.}
  \textbf{a}, A hue sweep measures decoded frequency and RGB.
  \textbf{b}, Ambiguous cues let observed motion select its paired motion--appearance mode.
  \textbf{c}, Long (65 frames) resists stronger conflicting cues than Short
  (8 frames); each rate uses 32 histories per cue and motion band.}
  \label{fig:solution-selection}
\end{figure*}

Appearance cues mainly select between the two training-supported frequency
bands, rather than continuously tuning speed. Around ambiguous hues, the
generated future is more likely to follow observed motion than at the red or
blue endpoints.

\resultlead{Observed motion completes the learned motion--appearance pair.} The
same model can also use observed motion to select the color paired with that
motion during training (Figure~\ref{fig:solution-selection}b). Under ambiguous
appearance cues, fast motion favors a blue future and slow motion a red one.
With strong color cues, the same model instead favors the frequency paired with
the observed color. These two behaviors show that the learned motion--appearance
association can be completed in either direction.

\resultlead{More visible motion shifts control toward the physical solution.} In
separately trained Short and Long models, motion-consistent futures persist
farther toward the conflicting red or blue endpoint under Long in both conflict
directions (Figure~\ref{fig:solution-selection}c). The generated future therefore
depends on the relative evidence supplied by appearance and observed motion.
Pendulum shows the same cue-dependent competition in another oscillatory system
(Appendix~\ref{app:pendulum-replication}).
We next test whether the motion-consistent alternative remains causally
accessible when the conflicting color cue controls generation.

\begin{figure*}[!ht]
  \centering
  \includegraphics[trim=0 11pt 0 1pt,clip,width=\textwidth]{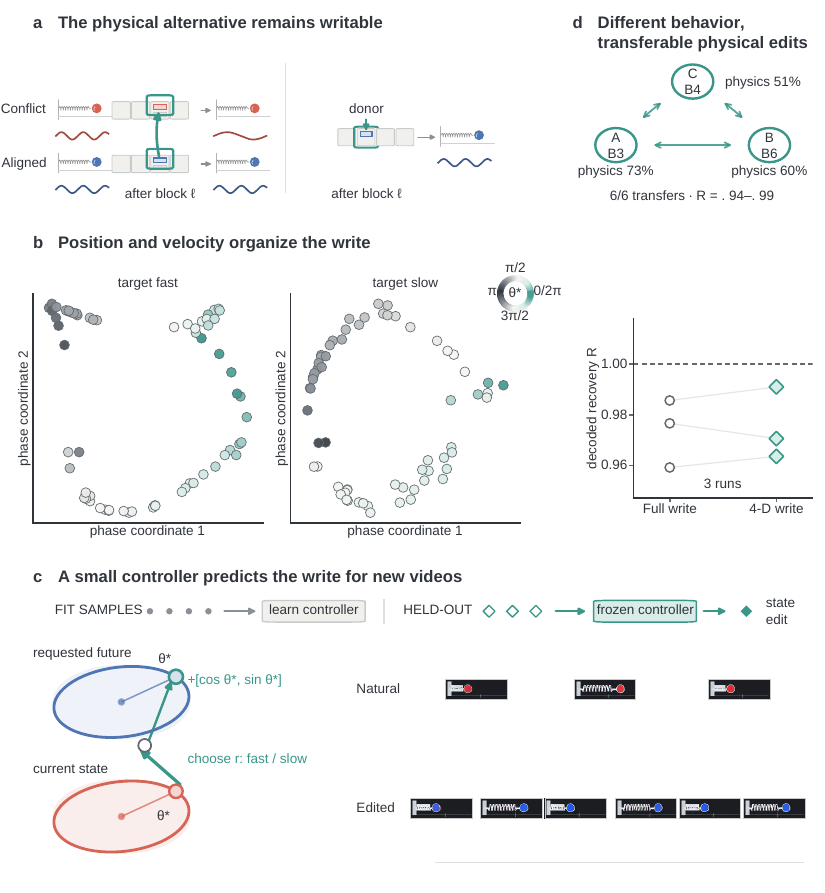}
  \caption{\textbf{A compact, physically structured activation edit restores
  the motion-consistent future.}
  \textbf{a}, A full edit redirects the naturally selected future in decoded
  video.
  \textbf{b}, Four coordinates retain nearly the full effect, with their
  variation organized by boundary position and velocity.
  \textbf{c}, A controller fitted on separate trajectories predicts held-out
  edits from target motion and boundary state.
  \textbf{d}, Natural physics-follow rates pool each run's $64\times11$ cue
  grid; all six directed transfers achieve held-out decoded recovery
  $R=.94$--$.99$.}
  \label{fig:causal-route}
\end{figure*}

\section{A Physically Structured Alternative Remains Causally Accessible}
\label{sec:causal-route}

We test causal accessibility in three independently trained Short models after 50K training steps. Each
strict aligned--conflict pair shares observed motion, prediction-boundary state,
and generation seed, and differs only in color. Both rollouts are valid: the
aligned input uses the training-paired color and follows observed motion,
whereas the conflict input uses the opposite color and follows appearance
(e.g., fast--blue versus fast--red; full criteria in
Appendix~\ref{app:experimental-details}). The prediction-boundary state is the
position and velocity at the final observed frame. Let $h_{A,i}$ and
$h_{C,i}$ denote their observed-frame token states after DiT block $\ell$
(with $i$ indexing an aligned--conflict pair).
Adding $d_i=h_{A,i}-h_{C,i}$ to the conflict run redirects its decoded future
(Figure~\ref{fig:causal-route}a).

Let $\hat\omega_{C,i}$ denote the fitted frequency of the natural conflict
rollout, $\hat\omega_{A,i}$ that of its corresponding aligned rollout, and
$\hat\omega_{\mathrm{edit},i}(\ell)$ the fitted frequency after intervention at
block $\ell$. We measure \textbf{normalized frequency recovery} by
\[
R_i^\omega(\ell)=
\frac{\hat\omega_{\mathrm{edit},i}(\ell)-\hat\omega_{C,i}}
     {\hat\omega_{A,i}-\hat\omega_{C,i}},
\]
so $R^\omega=0$ denotes the natural conflict frequency and $R^\omega=1$ the
corresponding aligned frequency. For the structure analysis, a layer scan
selects a late site where the write still redirects the decoded future; across
the three runs this site falls at blocks 3, 6, and 4. Only observed-frame tokens are
changed. This full edit does not reveal whether the model has learned a compact
physical representation. We therefore use PCA to look for low-dimensional
structure shared across edits.

\resultlead{Four coordinates retain nearly the full effect.} We fit an uncentered
PCA basis to paired differences from one set of trajectories. For each held-out
trajectory, we retain only the component of its paired difference along the
first four axes and test whether this low-rank edit preserves the effect of the
full difference. Across the three Short 50K runs, 87.8\% of held-out top-4 writes satisfy
$.75<R^\omega<1.25$, compared with 92.4\% for the full paired-difference edit.
Each run supports two rewrite directions (slow$\to$fast and fast$\to$slow), giving
six run--direction groups. This establishes a compact edit space, but each
held-out test still uses its own paired difference.

\resultlead{Boundary position and velocity organize the write.}
The Top-4 coordinates vary systematically with boundary phase: held-out edits
form smooth phase-ordered loops in both rewrite directions
(Figure~\ref{fig:causal-route}b).
For harmonic motion, $x=A\cos\theta$ and $v=-A\omega\sin\theta$, so the input
boundary phase
\[
\theta^*=\operatorname{atan2}(-v^*/\omega_{\rm true},x^*),
\]
compactly encodes normalized position and velocity: up to a common amplitude
scale, $\cos\theta^*$ is position-like and $\sin\theta^*$ is velocity-like after
frequency normalization. This organization suggests predicting an edit for a
new trajectory from its boundary state and requested target direction, without
the matched aligned activation. We fit a separate first-harmonic map for each
target family:
\[
\hat z^{(r)}_{1:4}(\theta^*)
=\beta^{(r)}_0+\beta^{(r)}_c\cos\theta^*
+\beta^{(r)}_s\sin\theta^*,
\qquad r\in\{\mathrm{fast},\mathrm{slow}\}.
\]
This model raises held-out coordinate $R^2$ to $.83$--$.88$, up
from $.51$--$.60$ for a direction-only model. For a held-out input, the frozen
controller predicts four coordinates from target direction and boundary state,
then reconstructs the activation edit through the frozen PCA basis. Across the
six run--direction groups, 85.9\% of these synthesized held-out writes satisfy
$.75<R^\omega<1.25$
(Figure~\ref{fig:causal-route}c).

Physical-state differences also place the two rewrite directions in a common
edit space (Appendix~\ref{app:state-parameterizations}).
The edits also rewrite associated appearance: slow$\to$fast shifts future color
toward blue. The same low-order organization recurs in Pendulum, where boundary
angle and angular velocity predict top-four coordinates, and in non-oscillatory
Free Fall, where a donor-free $[1,g_{\mathrm{target}}]$ controller reaches median held-out recovery
$R^g=.967$ across 64 decoded rollouts
(Appendix~\ref{app:pendulum-replication},~\ref{app:freefall-replication}).

\resultlead{Models with very different natural behavior can share transferable
physical corrections.} Despite identical architecture and training settings,
physics-follow rates span about 22 percentage points across the three training
seeds (Figure~\ref{fig:causal-route}d). This variation motivates testing whether
these models share a transferable physical edit space.
A scale and four-dimensional
orthogonal map fitted on separate
trajectories transfers controller-predicted edits across all six directed run
pairs, with near-full held-out decoded recovery
(Figure~\ref{fig:causal-route}d; Appendix~\ref{app:cross-run-transfer-controls}).
Their behavioral differences may therefore arise from how downstream computation
uses a shared physical solution. This shifts the question from whether the
physical solution exists to when it loses its ability to affect generation.

\section{Causal Writability: When Does the Future Become Hard to Change?}
\label{sec:writability}

To measure depthwise writability, we apply each matched full edit from
Section~\ref{sec:causal-route} at all 31 network sites. Early writes redirect the
shortcut future; after a narrow depth range, the edit stops changing decoded
motion. For a \textbf{strict-failure bank}---pairs whose aligned rollout
follows observed motion and whose conflict rollout follows color---and writing
$\Pr_i$ for the fraction over trajectories $i$, define
\begingroup
\[
W^\omega(\ell)
=\Pr_i\!\left[\mathrm{valid}\ \land\ .75<R_i^\omega(\ell)<1.25\right],
\qquad
D^\omega=\sum_{\ell\in\mathcal L}W^\omega(\ell),
\]
over the common 31-site scan $\mathcal L$ (one site before block 0 and one after
each of the 30 blocks). $W^\omega(\ell)$ is the fraction of strict failures that
can still be redirected at site $\ell$. $D^\omega$ is the expected number of
writable sites out of 31; a larger value means that the same write remains
effective deeper into the network. For one trajectory,
\[
D_i^\omega
=\sum_{\ell\in\mathcal L}
\mathbf 1\!\left[\mathrm{valid}\ \land\ .75<R_i^\omega(\ell)<1.25\right].
\]
\endgroup
The motion-consistent future is \textbf{causally writable} while the tested write
changes the decoded motion; closure marks \textbf{commitment} for the write. Individual
trajectories close at different sites, so averaging broadens the population
profile. In Pendulum, more observed motion delays closure in both directions;
Free Fall loses near-full direct writability after block 1
(Appendix~\ref{app:pendulum-replication},~\ref{app:freefall-replication}).

\begin{figure}[!t]
  \centering
  \includegraphics[width=.893\textwidth]{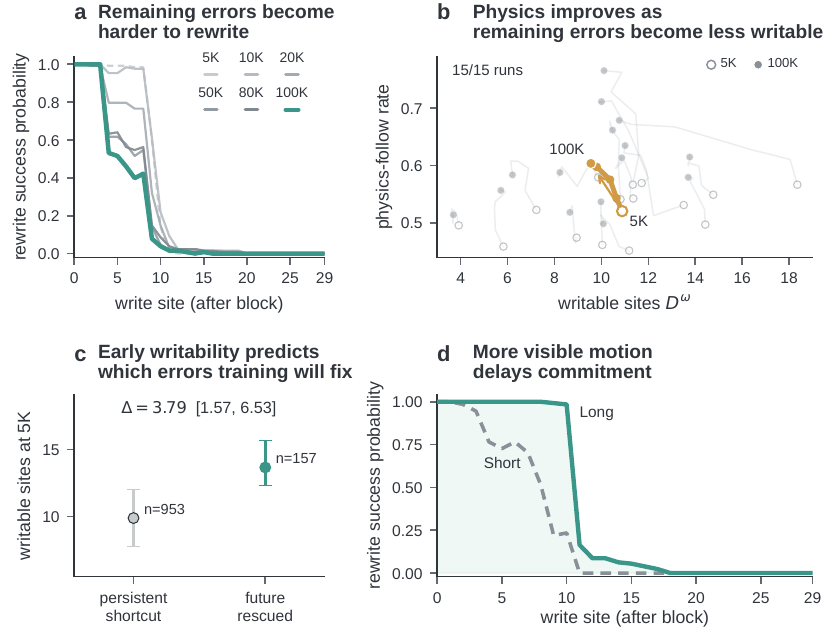}
  \caption{\textbf{Causal writability closes with depth and predicts which
  early failures training later corrects.}
  \textbf{a}, Layerwise recovery closes earlier over training.
  \textbf{b}, Across 15 seeds, mean full-grid ($64\times11$) physics-follow rate
  rises while remaining errors become less writable.
  \textbf{c}, Later-corrected errors are more writable at 5K.
  \textbf{d}, Long (65 frames) delays closure relative to Short (8 frames).}
  \label{fig:writability}
\end{figure}

\resultlead{Training corrects some errors while consolidating the rest.} Across
15 Short runs, physics-following behavior rises from $.52$ at 5K to $.60$ at
100K, while mean writability among remaining shortcut errors falls from $10.88$
to $9.55$ sites (Figure~\ref{fig:writability}b). A classifier can improve held-out
accuracy while becoming more confidently wrong on some remaining examples,
increasing their cross-entropy loss. We therefore ask whether early writability
distinguishes the shortcut errors that training later corrects from those that
persist. See Appendix~\ref{app:writability-robustness} for
same-checkpoint comparisons.

\resultlead{Early writability predicts which errors training later corrects.} We
test this prediction using trajectories that all follow the shortcut at 5K.
Future-rescued trajectories follow observed motion at both 80K and 100K, whereas
persistent failures follow the shortcut at all six checkpoints (5K, 10K, 20K,
50K, 80K, and 100K). Future-rescued errors are writable at $3.79$ more sites on
average (95\% CI $[1.57,6.53]$; Figure~\ref{fig:writability}c). Early writability
therefore separates the two outcomes before their generated behavior diverges
(Appendix~\ref{app:writability-robustness}).

The link between early writability and later behavior raises a second question:
can stronger support for the physical future keep it writable deeper into the network?
We test two ways of providing that support: more observed motion and learning
dynamics before introducing the appearance shortcut.

\resultlead{More visible motion delays commitment.} In all three matched-seed
comparisons at 50K, Long remains writable to a
later block than Short (Figure~\ref{fig:writability}d).

\resultlead{Training order moves the commitment boundary.}
Longer histories strengthen the motion evidence available at inference.
We next ask whether learning motion first has a similar effect, comparing
two adaptation paths for the same pretrained Wan 1.3B video DiT.
Direct adaptation learns the biased red--slow/blue--fast task immediately.
Neutral-first adaptation first learns the same motion from achromatic videos,
where appearance carries no frequency information, and then receives the same
biased data.

\begin{figure}[!htbp]
  \centering
  \includegraphics[trim=0 16pt 0 10pt,clip,width=\textwidth]{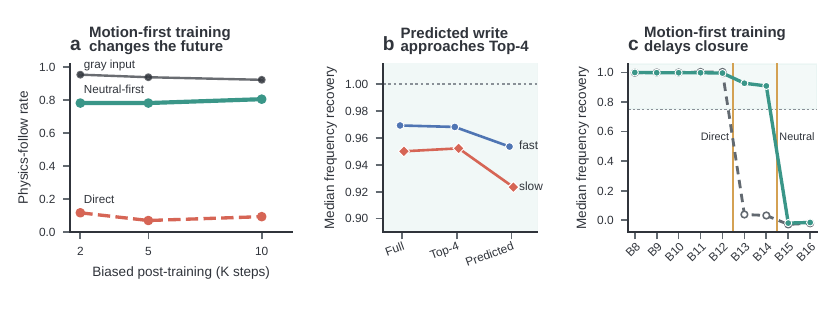}
  \caption{\textbf{In a pretrained video DiT, learning cue-independent dynamics
  first improves physical behavior and delays commitment.} \textbf{a}, Direct
  and Neutral-first conflict-endpoint rates and the matched gray-input rate
  each use the same 128 held-out histories.
  \textbf{b}, The donor-free
  controller approaches the full-edit and Top-4 references.
  \textbf{c}, Direct and Neutral-first adaptation both close sharply, but
  between B12--B13 and B14--B15, respectively.}
\label{fig:pretrained-curriculum}
\end{figure}

Training order changes the generated future. On a fixed population of
128 trajectories, conflict physics-follow at 10K is $80.5\%$ after Neutral-first
adaptation and $9.4\%$ after Direct adaptation. Aligned performance is similar
(Figure~\ref{fig:pretrained-curriculum}a). Before
comparing closure, we verify that the compact edit remains effective in
Wan. A controller fitted on separate trajectories approaches the Top-4
reference in both write directions (Figure~\ref{fig:pretrained-curriculum}b).

Learning cue-independent motion first delays closure. On the same
persistent-failure identities, Neutral-first adaptation retains writability
deeper than Direct adaptation
(Figure~\ref{fig:pretrained-curriculum}c). On fixed identities, access can close
and later reopen across nearby training checkpoints. Commitment therefore marks
a depth boundary within a given forward pass; subsequent training can move or
reopen it
(Appendix~\ref{app:pretrained-curriculum}).

Longer histories and motion-first learning thus provide two ways to preserve
access to the physical future. They leave open what stops that future from
controlling generation when direct writing finally fails.

\section{Attention Writes the Selected Future into Future Tokens}
\label{sec:downstream}
When direct writing stops changing the video, two explanations remain. The
signal for the alternative motion may have disappeared, or it may still be
present after the selected motion has already been written into future-frame
tokens.

\begin{figure}[!t]
  \centering
  \includegraphics[width=\textwidth]{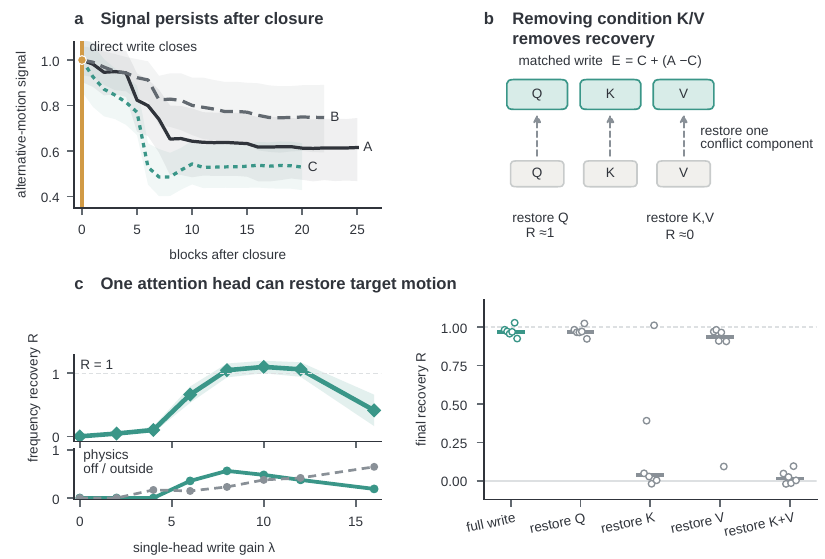}
  \caption{\textbf{The alternative-motion signal survives closure, and
  self-attention determines whether it changes the video.}
  \textbf{a}, The signal remains measurable after observed-frame writing stops
  changing decoded motion.
  \textbf{b}, Restoring conflict K+V, but not Q, removes frequency recovery.
  \textbf{c}, On 48 held-out fast-motion/red-cue strict failures, moderate
  V-head amplification restores target motion; excessive gain leaves the
  supported modes.}
  \label{fig:downstream}
\end{figure}

\resultlead{The alternative-motion signal survives closure.} We first apply a write
at an earlier writable block. At each subsequent block, we compare the edited
future-frame activations with the natural conflict run and project their
difference onto the aligned-minus-conflict direction. We call this projection
the alternative-motion signal. A value of 1 indicates the full paired
difference along that direction; 0 indicates the unedited conflict response. On
held-out fast-target examples, the signal
remains measurable in all three runs from the first non-writable block through
the final block, even after observed-frame writes stop changing the generated
frequency (Figure~\ref{fig:downstream}a).

We next localize the pathway through which the physical direction causally
affects generation, starting from the full edit $E=C+(A-C)$
(where $C$, $A$, and $E$ denote the conflict, aligned,
and edited activations, respectively). We first replace either the
post-attention future-frame state or the subsequent MLP output with its
natural-conflict counterpart. We then decompose self-attention by restoring its
natural-conflict Q, K, or V
components one at a time. Restoring Q tests the queries issued by future-frame
tokens, while restoring K or V tests the information made available by
observed-frame tokens.

\resultlead{Observed-frame K/V mediates the motion write into future-frame tokens.}
Restoring the natural-conflict post-attention future-frame state removes
frequency recovery, whereas restoring the following MLP output preserves
frequency recovery.
This localizes the write to self-attention. Within self-attention, restoring Q
has little effect, while jointly restoring observed-frame K+V removes recovery
in all six run--direction groups (Figure~\ref{fig:downstream}b). Future-frame
queries therefore obtain the motion-controlling information from observed-frame
K/V.

We next test whether amplifying this attention write can restore control.
Head-level responsibility differs across runs: the effect is concentrated in
one K head, distributed across K/V, or localized to a selective V head
(Appendix~\ref{app:implementation-multiplicity}). In Run C, we amplify
only the observed-frame V of one of the nine heads:
\[
V_C^{(8)}+\lambda\,\Delta V^{(8)},
\qquad
\Delta V^{(8)}=V_M^{(8)}-V_C^{(8)},
\]
where $M$ is the state produced by the frozen controller from
Section~\ref{sec:causal-route} and
$\lambda$ scales the strength of the physical-alternative direction.

\resultlead{Moderate amplification allows the alternative motion to regain control.} At
$\lambda=8$, the physics-following rate
peaks at $56.3\%$ across 48 held-out strict failures
(Figure~\ref{fig:downstream}c). Higher gain overshoots: at $\lambda=16$, all
outputs remain valid, but most are off-family.
In this model, the alternative-motion signal remains present,
but its unamplified attention write is too weak to control the video. We also
examine when the alternative-motion write becomes effective during iterative
generation. Early-only flow-matching writes produce an internal response without
decoded recovery, whereas late- or all-call writes restore the motion-consistent
future. Effective control therefore depends on the temporal allocation of the
write, not only its location within the network
(Appendix Figure~\ref{fig:fm-time-decoded-rescue}).

\section{Discussion and Limitations}

Predictive underspecification and causal writability answer different questions.
The former asks which solutions fit the training data; the latter asks which can
still reach the output as inference proceeds. This parallels the
control-theoretic distinction between observability and controllability
\citep{kalman1960general}: a physical solution may remain detectable after it
loses control over generation. Commitment marks this loss of causal access, not
the loss of physical information.

Our mechanistic analysis uses controlled synthetic dynamics, where exact
counterfactual matching and quantitative decoded-video evaluation are possible.
The learned edits follow the coupled appearance--dynamics modes in training, and
closure is specific to the observed-frame edit, sites, and checkpoints tested.
Independent runs distribute the observed-to-future attention write across
different K/V components and heads. Pendulum extends the analysis to another
oscillator and renderer; Free Fall extends it to non-oscillatory dynamics with a
specified-gravity controller. In a pretrained 1.3B Wan video DiT, we likewise
observe an executable physical
edit and a clear closure boundary, showing that the acquisition--control
distinction is not confined to models trained from scratch. Whether naturally
occurring shortcuts in broadly pretrained world models exhibit the same point
of no return remains open.

\section{Related Work}

\paragraph{Shortcut learning and predictive underspecification.}
Training data can support multiple predictive rules that agree on the training
distribution and diverge only under shift
\citep{geirhos2020shortcut,damour2022underspecification}.
In controlled mechanics videos, \citet{kang2025physical} find case-based
generalization with an attribute hierarchy that places color above velocity.
Our cue sweep also reveals the reverse relation: observed motion can select its
training-associated color.

\paragraph{Physics representations and control in video models.}
Physics-related structure is decodable in video encoders
\citep{joseph2026interpreting} and inverted diffusion trajectories
\citep{esmati2026invisible}; concept activation vectors steer VideoMAE's
plausibility judgments \citep{alam2026causal}. Generation improves through
relational alignment \citep{zhang2025videorepa} or inference-time rewards
\citep{yuan2026physicsalignment}. We test whether physical-state-based activation
edits change motion in fully decoded video, distinguishing the presence of
physical information from its ability to control generation.

\paragraph{Causal intervention, commitment, and cross-model transfer.}
Activation patching probes internal mechanisms
\citep{meng2022locating,zhang2024patching}, but subspace interventions can
mislead causal attribution \citep{makelov2024subspace}. Manifold steering links
geometry to behavior, including video physics \citep{wurgaft2026manifold}.
\citet{plattner2026circuits} localize and rescue 3D failure through an
early-denoising cross-attention write. Our question is whether a physically
specified alternative remains causally usable after the model has selected a
different future.
Stitching \citep{bansal2021stitching} and permutation alignment
\citep{ainsworth2023git} compare networks, but stitching does not establish
shared information \citep{smith2025alignment}. Cross-model steering transfers
interventions \citep{oozeer2025activationtransfer,poppi2026crossmodel}; we test
held-out physical corrections across models with different natural behavior.

\section*{Reproducibility statement}

Appendix~\ref{app:experimental-details} documents the model, data generation,
decoded evaluator, statistical units, strict-pair construction, and intervention
protocol. We will release training and evaluation code, configurations, data
generators, random seeds, checkpoint metadata and hashes, frozen evaluation
manifests, frozen aligned--conflict pair identities and input frames, figure
builders, and audit scripts. The manifests
record the sample identities used in each comparison and link quantitative
figures to their source artifacts. A release script reruns the evidence checks,
verifies provenance, rebuilds the mechanism figures, and compiles the
manuscript. The release will include all hyperparameters and
experiment-specific settings needed to reproduce the reported results.

\section*{AI use statement}

The authors used generative AI assistants, including OpenAI ChatGPT and
Codex-based coding agents, to help refine hypotheses and experimental designs;
implement and refactor code; run and monitor experiments; construct audits;
analyze results; design figures; search the literature; and draft, translate,
and edit the manuscript. The authors formulated the research questions, made
all final methodological and interpretive decisions, approved each experiment,
supervised its execution, inspected raw outputs, and reviewed the
code, citations, figures, and claims in the submission. All reported
measurements come from the stated deterministic evaluators and intervention
pipelines, not language-model judgments. The authors reviewed all AI-assisted
material and take responsibility for the final paper.

\bibliography{refs}
\bibliographystyle{iclr2027_conference}
\appendix
\section{Experimental Details}
\label{app:experimental-details}

\paragraph{Model and training.}
The Spring model is a no-text latent flow-matching transformer with
488,158,912 trainable parameters, excluding its frozen VAE. Its denoiser has 30
bidirectional DiT blocks, hidden width 1152, feed-forward width 4608, nine
attention heads of width 128, and a $1\times2\times2$ latent patch size. A frozen
Wan2.1 VAE maps each 129-frame video to a $33\times16\times16$ latent grid with
16 channels: 17 conditioning frames followed by 16 target frames. Spatial
patching gives 1,088 condition and 1,024 target tokens.
The two history regimes use the same architecture, token count, prediction
boundary, and optimizer; they differ only in the visible physical history.
Table~\ref{tab:experimental-contract} gives the remaining shared contract.

\begin{table*}[!htbp]
  \centering
  \caption{\textbf{Spring experimental contract.} Long and Short use the same
  video span and latent layout; Short replaces early motion with the fixed
  background before VAE encoding.}
  \label{tab:experimental-contract}
  \begin{tabular}{p{0.27\textwidth}p{0.66\textwidth}}
    \toprule
    Component & Setting \\
    \midrule
    Training support & 2,048 videos per regime: 1,024 red--slow and 1,024 blue--fast; no conflicting color--frequency combinations \\
    Physical variables & slow $\omega\sim U[2.2,3.0]$ rad/s; fast $\omega\sim U[5.2,6.4]$ rad/s; normalized horizontal-displacement amplitude $\sim U[0.10,0.17]$; phase $\sim U[0,2\pi)$ \\
    Video and boundary & $128\times128$, 20 fps; frames 0--64 are conditioning frames and frames 65--128 are generated \\
    Long / Short history & Long exposes real frames 0--64; Short replaces frames 0--56 by RGB $(28,30,34)$ and exposes real motion in frames 57--64; replacement occurs before VAE encoding \\
    Renderer & equilibrium $x=0.58$, anchor $x=0.16$, center $y=0.50$, mass radius 7 px, 10 spring coils; red RGB $(235,48,48)$ and blue RGB $(48,96,235)$. Oscillator position $x$, amplitude, and boundary position $x^*$ are expressed as normalized horizontal displacements from equilibrium; $v^*$ is the corresponding displacement velocity per second. A displacement $x$ is rendered at horizontal pixel coordinate $(0.58+x)(W-1)$ \\
    Latent sequence & 33 latent frames and 2,112 tokens: 17 condition frames (1,088 tokens) and 16 target frames (1,024 tokens); only target frames enter the loss \\
    Optimization & AdamW with PyTorch ConstantLR, batch size 32, gradient accumulation 1, base learning rate $2\times10^{-4}$, weight decay $0.01$, no EMA \\
    Training schedule & primary mechanism analyses use the 50K checkpoints; the 15-run longitudinal runs train to 100K and are evaluated at 5K, 10K, 20K, 50K, 80K, and 100K \\
    Sampling & 20 flow-matching calls, scheduler shift 5.0, denoising strength 1.0, and fixed per-trajectory generation seeds \\
    \bottomrule
  \end{tabular}
\end{table*}

\paragraph{Decoded-video evaluation.}
Evaluation uses generated pixels.
A deterministic detector tracks the mass center in every future frame without
reading the input hue label. Validity requires detection in at least 90\% of
frames, no missing run longer than three frames, adjacent and boundary jumps at
most 12 px, vertical deviation at most 7 px, multi-component rate at most
0.10, and valid area and image bounds. For valid tracks, detected horizontal
pixel coordinates are converted to normalized displacement from equilibrium. We
fit the detected future-frame positions at their original time points, without
interpolating missing detections, using
\[
x(t)=a\cos(\hat\omega t)+b\sin(\hat\omega t)+c
\]
over a 1,601-point grid with $\hat\omega\in[1.21,8.64]$ rad/s. Fits with RMSE at
most $0.035$ are retained; the RMSE is measured in the normalized displacement
coordinate. Frequency-band membership uses a $0.005$ rad/s tolerance at the slow
and fast band boundaries. We
measure future appearance from the mean RGB of the detected mass pixels across
valid future frames---not by averaging entire video frames---and project it onto
the red-to-blue color axis. The 11 cues linearly interpolate between RGB
$(235,48,48)$ and $(48,96,235)$. A valid
rollout is labeled physics-following when $\hat\omega$ lies in the
history-consistent band, shortcut-following when it lies in the cue-associated
opposite band, compromise when it lies in the unsupported gap, and
off-family/invalid otherwise. The frozen 64-history $\times$ 11-hue
response grid keeps trajectory identity, physical parameters, renderer, and
generation seed fixed across hue. Consequently, hue variants are repeated
measurements of 64 histories, not 704 independent samples.

\paragraph{Strict aligned--conflict pairs.}
A strict pair consists of an aligned rollout that is valid and in its true
frequency band and a conflict rollout that is valid and follows the opposite
shortcut band. The pair shares trajectory, frequency, amplitude, phase,
boundary state $(x^*,v^*)$, renderer, and generation seed; only mass color
changes. The route and cross-run analyses use a shared 256-trajectory bank for
the three 50K Short runs: 128 fast-target and 128 slow-target pairs, split
into 128 fit and 128 held-out pairs balanced by direction. Every pair satisfies
the strict criteria in all three runs and has
$|\hat\omega_A-\hat\omega_C|\ge2.0$. Writability uses each
seed--checkpoint's local bank of natural strict failures, because that estimand
asks which errors remain at that point in training. The persistent-error control
freezes trajectory identities across checkpoints to test bank-composition
effects directly.

\paragraph{Intervention operators.}
Layer localization and writability replace the condition state,
\[
h^C_{\ell,\mathrm{cond}}\leftarrow h^A_{\ell,\mathrm{cond}},
\]
whereas route analyses add a full, projected, transferred, or predicted edit,
\[
h^C_{\ell,\mathrm{cond}}
\leftarrow h^C_{\ell,\mathrm{cond}}+\tilde d_\ell,
\qquad d_\ell=h^A_{\ell,\mathrm{cond}}-h^C_{\ell,\mathrm{cond}}.
\]
Both operators modify only the 1,088-token condition prefix at one site, hit
that site once on each of 20 flow-matching calls, and leave the 1,024-token
target suffix exactly unchanged at injection; the suffix then evolves normally
through subsequent computation. Flow-matching-window and component-restoration experiments
state their deviations explicitly. Layer scans cover 31 sites: the pre-block
residual and after-B0 through after-B29. Route localization chooses the latest
after-block at which at least 75\% of edits in each direction are valid and
satisfy $0.75<R_i^\omega<1.25$; this functional site is distinct from
$D^\omega$, which summarizes writability over the full scan.

We call an analysis \emph{fit-only} when every projection, scale, coefficient,
and model-selection choice is fixed on the fit split before any held-out
trajectory is evaluated.

\paragraph{Statistical units and provenance.}
Figure~\ref{fig:solution-selection} treats the 64 physical histories (32 per
frequency band) as units and the 11 hues as repeated measurements. Route,
controller, and transfer fits use trajectory-disjoint held-out sets. The
cross-solution analysis resamples complete six-checkpoint seed trajectories;
the future-fate analysis first resamples seeds and then trajectories within
seed $\times$ direction $\times$ fate strata. Exact checkpoint, bank, split,
bootstrap seed, validity record, and source-artifact hash are frozen, where
applicable, in figure-specific machine-readable manifests.

\section{Supplementary Results}

Throughout this section, Runs A, B, and C denote the independently trained
Short models with seeds 3407, 3408, and 3409, respectively.

\subsection{Behavioral robustness and matched cue reversal}
\label{app:endpoint-purple}
\label{app:multiseed-behavior}

Table~\ref{tab:endpoint-purple} conditions on trajectories for which the strongly
conflicting endpoint cue already produces the shortcut-associated frequency: red for
a true-fast history and blue for a true-slow history. It then changes only the cue to
purple ($u=0.5$) for the same physical trajectory and generation seed. Thus the table
measures a within-trajectory change in the naturally selected decoded continuation,
not accuracy on two unrelated sample sets.

\begin{table*}[!htbp]
  \centering
  \caption{\textbf{Endpoint shortcut failures often return toward the
  history-consistent continuation at the ambiguous cue.} Purple outcomes are
  counts within each endpoint shortcut cohort.}
  \label{tab:endpoint-purple}
  \setlength{\tabcolsep}{3pt}
  \begin{tabular}{lcrrrr}
    \toprule
    Run / motion & \shortstack{Endpoint\\shortcut} & \shortstack{Purple\\physics} & \shortstack{Purple\\shortcut} & \shortstack{Purple\\compromise} & \shortstack{Purple\\off/invalid} \\
    \midrule
    A / fast & $10/32$ & 8 & 0 & 1 & 1 \\
    A / slow & $16/32$ & 15 & 0 & 1 & 0 \\
    B / fast & $16/32$ & 9 & 1 & 5 & 1 \\
    B / slow & $8/32$ & 8 & 0 & 0 & 0 \\
    C / fast & $8/32$ & 1 & 0 & 5 & 2 \\
    C / slow & $9/32$ & 9 & 0 & 0 & 0 \\
    \bottomrule
  \end{tabular}
\end{table*}

The categorical reversal is solution- and direction-dependent, especially for
the fast-history branch of Run C. Nevertheless, over the endpoint-shortcut
cohort, every run--direction group shifts both frequency and future color toward
the history-associated joint mode on average: the mean frequency shift ranges
from $1.68$ to $2.76$ rad/s and the normalized color shift from $.50$ to $.71$.

\begin{figure}[!htbp]
  \centering
  \includegraphics[width=\textwidth]{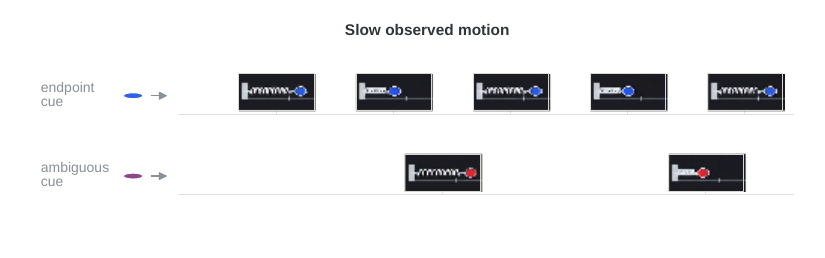}
  \caption{\textbf{The reverse matched example shows the same coupled switch.}
  With slow observed motion and the future window fixed, changing only the cue
  from the conflicting blue endpoint to the ambiguous cue redirects decoded
  motion toward the slow band and future appearance toward red. Frames are the
  measured turning points from the same decoded future window.}
  \label{fig:slow-matched-example}
\end{figure}

\begin{figure}[!t]
  \centering
  \includegraphics[width=\textwidth]{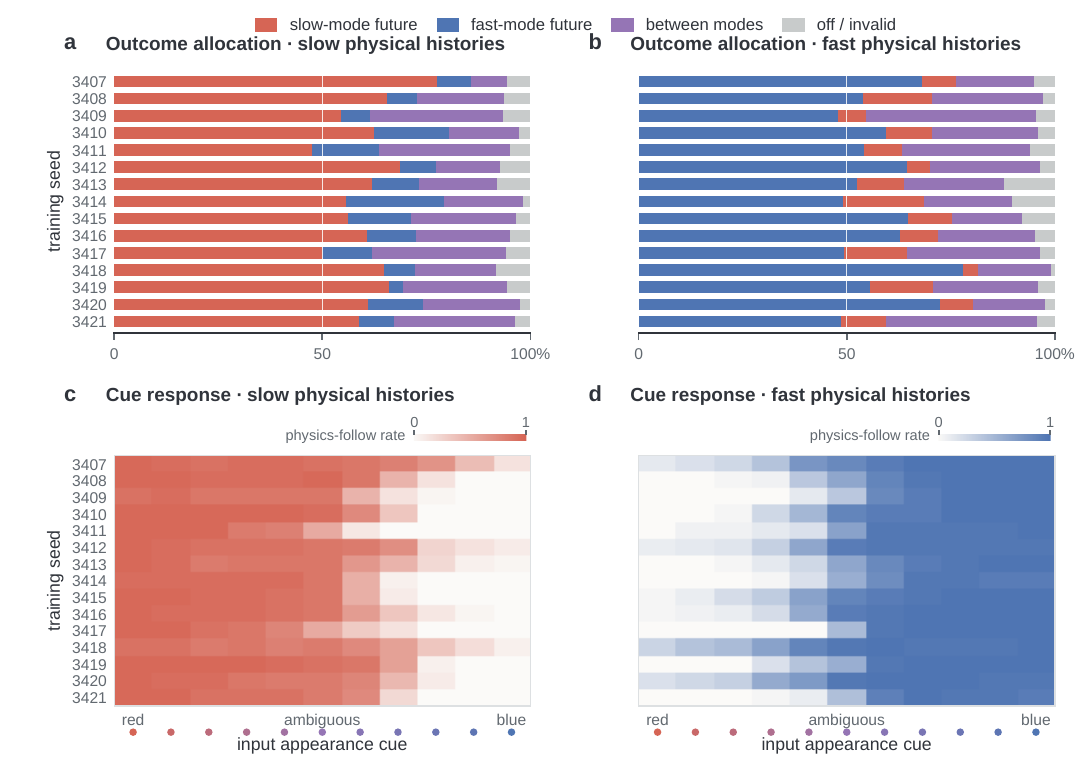}
  \caption{\textbf{Evidence-dependent joint-mode selection recurs across 15
  independently trained models.} \textbf{a,b}, Outcome allocation over all 352
  hue-conditioned rollouts in each physical band. Colors denote decoded motion
  modes, not correctness: coral is slow, blue is fast, purple lies between the
  supported bands, and gray is off-family or invalid. \textbf{c,d},
  Physics-follow rate for each seed and cue; every cell averages 32
  physical histories.}
  \label{fig:multiseed-behavior}
\end{figure}

Figure~\ref{fig:multiseed-behavior} expands the decoded-video evaluation to all
15 Short runs at 50K. Each run uses the same 64 physical histories, 11
hues, and deterministic evaluator; all 10,560 rollouts are valid. The aggregate
allocation varies substantially across runs, particularly in how often an
unsupported compromise is realized. The cue-response panels nevertheless show
the same evidence-dependent pattern: slow histories are most often preserved
near red cues, fast histories near blue cues, with solution-specific transition
widths around the ambiguous region.

\FloatBarrier

\subsection{Pretrained Wan replication of the executable route and sharp closure}
\label{app:pretrained-controller}

We repeat the route analysis in a pretrained Wan 1.3B video DiT adapted to the
biased Spring task. A direction-specific first-harmonic controller is fit on
128 trajectories and frozen before evaluation on 128 disjoint receivers. The
held-out coordinate fits reach $R^2=.858/.771$ for fast/slow targets. In decoded
rollouts, the donor-free controller attains median recovery
$R^\omega=.953/.923$, close to the Top-4 ceiling of $.968/.952$;
neither controller fitting nor held-out intervention accesses the held-out
aligned activation. Top-4 oracle edits are rescaled to the receiver's full-edit
norm; phase-predicted edits are not.

The same model family also exhibits localized closure. On the same 17
slow-target persistent failures, direct adaptation loses median recovery between
B12 and B13, whereas neutral-first adaptation retains it until B14 and loses it
between B14 and B15. Thus the training path shifts the boundary without turning
the loss of direct condition-side control into a gradual network-wide decay.

\begin{figure}[!htbp]
  \centering
  \includegraphics[width=\textwidth]{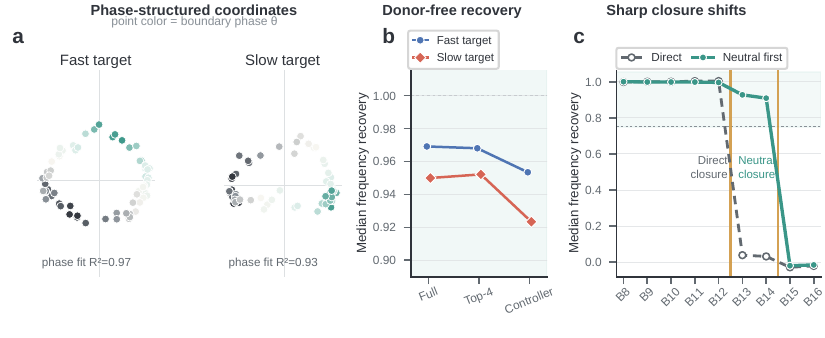}
  \caption{\textbf{A pretrained video DiT reproduces the executable route and
  localized commitment boundary.} \textbf{a}, Held-out top-four coordinates
  vary smoothly with boundary phase in both rewrite directions ($n=64$ each;
  the displayed state-varying coordinate pairs are centered for visualization).
  \textbf{b}, The Top-4 projection retains the full-edit
  recovery, and the fit-only phase controller approaches that ceiling without a
  held-out donor. \textbf{c}, Median recovery on the same persistent-failure
  identities drops across one adjacent-block transition under both training
  paths, while neutral-first adaptation shifts the transition later. The amber
  lines mark the two observed closure intervals; the shaded region denotes
  $.75<R^\omega<1.25$.}
  \label{fig:pretrained-wan-route-boundary}
\end{figure}

\FloatBarrier

\subsection{Pretrained neutral-to-biased curriculum}
\label{app:pretrained-curriculum}

We initialize a Wan 1.3B video DiT in two ways. The Direct path adapts the
pretrained checkpoint directly on the red--slow/blue--fast Spring data. The
Neutral path first trains for 10K updates on 2,048 achromatic Spring videos,
with slow and fast dynamics equally frequent and all masses rendered RGB
$(92,92,92)$, then starts the same biased stage from a fresh optimizer. The
biased-stage data order and random streams are matched between paths. Gray and
biased training manifests have disjoint trajectory identities.

Population results use a frozen evaluation set generated before model
evaluation: 128 physical histories (64 per frequency band), each rendered at 11
red-to-blue hues and one gray appearance. Its base seeds are disjoint from both
training manifests, the earlier development hue set, and the mechanism cohort.
Table~\ref{tab:pretrained-heldout} reports endpoint behavior. Every
biased-checkpoint endpoint and gray rollout is evaluator-valid, so the nonzero
off-family entries are valid tracks whose fitted frequencies lie outside the
supported outcome categories.

\begin{table}[!htbp]
  \centering
  \caption{\textbf{Neutral adaptation redirects the solution reached under the
  same subsequent biased data.} Columns report decoded conflict outcomes on the
  same 128 held-out histories. The paired interval resamples physical histories.}
  \label{tab:pretrained-heldout}
  \setlength{\tabcolsep}{2pt}
  \begin{tabular}{llrrrrrr}
    \toprule
    \shortstack{Biased\\step} & Path & \shortstack{Aligned\\physics} & \shortstack{Conflict\\physics} & Shortcut & Compromise & \shortstack{Off-\\family} & \shortstack{Neutral$-$Direct\\physics} \\
    \midrule
    2K & Direct  & 83.6 & 11.7 & 61.7 & 16.4 & 10.2 & \\
       & Neutral & 93.8 & 78.1 & 12.5 & 7.8  & 1.6  & 66.4 [57.8,74.2] \\
    5K & Direct  & 93.0 & 7.0  & 65.6 & 23.4 & 3.9  & \\
       & Neutral & 96.1 & 78.1 & 7.0  & 14.8 & 0.0  & 71.1 [63.3,78.9] \\
    10K& Direct  & 96.1 & 9.4  & 58.6 & 28.1 & 3.9  & \\
       & Neutral & 93.8 & 80.5 & 7.8  & 11.7 & 0.0  & 71.1 [63.3,78.9] \\
    \bottomrule
  \end{tabular}
\end{table}

The identical 5K and 10K paired intervals are not duplicated estimates. Both
paired difference vectors contain 91 improvements and 37 ties, although 20
trajectory identities change outcome between the two checkpoints; their
trajectory-bootstrap distributions therefore coincide.

At the achromatic 10K starting checkpoint, the median matched red--blue
endpoint frequency difference is $.056$ rad/s and opposite-band shortcut
outcomes are absent at the nominal conflict endpoint. The checkpoint is an
imperfect Spring generator, especially for fast histories, but hue does not yet
control the generated frequency systematically. Within the Neutral biased path,
gray inputs for the same histories remain more physics-following than colored
conflicts by 17.2, 15.6, and 11.7 percentage points at 2K, 5K, and 10K,
respectively.

\begin{figure}[!t]
  \centering
  \includegraphics[width=\textwidth]{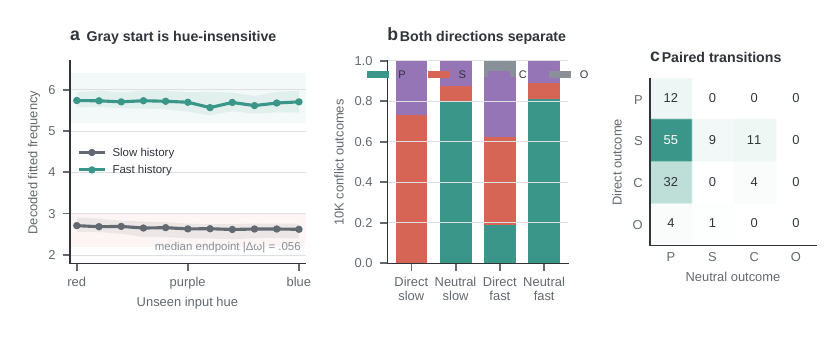}
  \caption{\textbf{A cue-independent starting solution redirects biased
  post-training on a frozen held-out population.} \textbf{a}, Before biased
  training, median decoded frequency changes little over unseen hue; shading is
  the interquartile range across histories. \textbf{b}, The 10K conflict
  difference occurs in both physical directions. \textbf{c}, Counts for the
  same 128 histories show that
  most Direct shortcut/compromise outcomes become physical under Neutral
  initialization. The post-training population curves already shown in
  Figure~\ref{fig:pretrained-curriculum}a are not repeated here.}
  \label{fig:pretrained-behavior-appendix}
\end{figure}

The event-centered causal analysis uses a separate seed-disjoint bank enriched
for shortcut-susceptible histories and therefore does not estimate population
failure prevalence. At 2K, 17 slow-direction failures already present at 1K
have median deepest writable block B12, whereas ten failures newly observed at
2K have median B15 (difference three blocks, trajectory-bootstrap 95\% CI
$[2,4]$). Among all 27 slow strict failures at 2K, those still shortcut-selected
at 3K have median boundary B12, versus B14.5 for non-persistent failures
(difference 2.5 blocks, 95\% CI $[2,3]$).

The 2K depth ordering remains strong for behavior measured at 4K and 5K but
attenuates by 10K, supporting a near-term stabilization signal whose predictive
power weakens over longer horizons. On 11 fixed identities, recovery at B13/B14 also closes
and reopens over nearby checkpoints while B12 remains writable and B15 remains
typically closed. This exploratory fixed-identity result makes commitment a
checkpoint-local computational property, not an absorbing state across further
optimization.

\begin{figure}[!t]
  \centering
  \includegraphics[width=\textwidth]{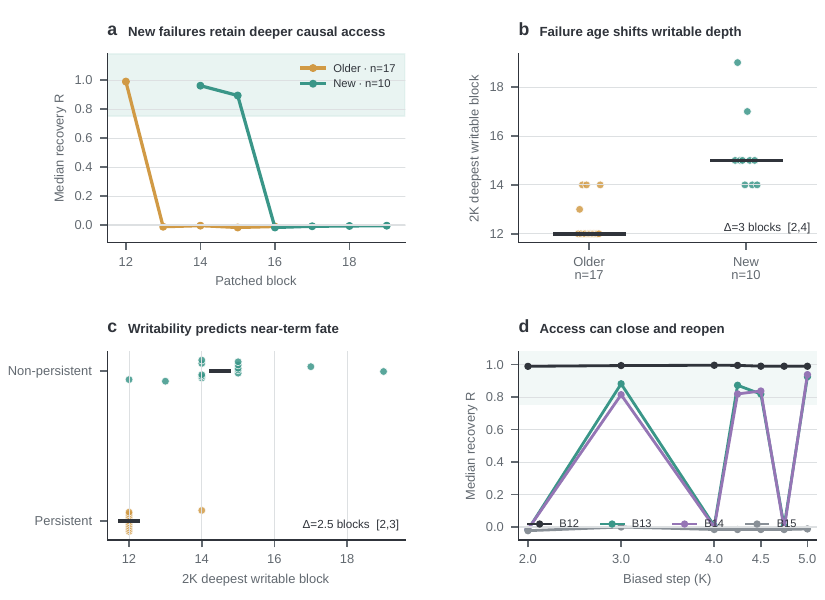}
  \caption{\textbf{Behavioral failure precedes stable writability closure in the
  Neutral path.} \textbf{a}, At 2K, newly observed failures retain effective
  edits about three blocks deeper than failures already present at 1K.
  \textbf{b}, The same comparison is summarized per trajectory by its deepest
  writable block. \textbf{c}, Among the same 2K failures, deeper writability
  predicts departure from the shortcut mode at 3K. \textbf{d}, On fixed
  persistent identities, downstream B13/B14 access can close and reopen across
  nearby checkpoints; curves show cohort medians.}
  \label{fig:pretrained-causal-timing-appendix}
\end{figure}

\FloatBarrier

\subsection{Boundary-phase controller ablations}
\label{app:boundary-phase-controller}

Across the six Spring run--direction groups in the main-text comparison,
median normalized recovery ranges from $.910$ to $1.012$ for the Top-4
paired-difference edit and from $.910$ to $1.009$ for the predicted
boundary-state controller.

The two-dimensional view in Figure~\ref{fig:causal-route}b is a fit-only
phase-aligned projection inside the frozen top-four route. After fitting the
joint raw, uncentered PCA on 128 fit differences, we fit within each target
direction
\[
z(\theta^*)=\mu+a\cos\theta^*+b\sin\theta^*
\]
in four-dimensional route coordinates and orthonormalize the span of $[a,b]$.
This frozen plane is then applied to 64 disjoint held-out pairs per direction.
For the displayed run, held-out phase-plane $R^2$ is $.935/.881$; the
corresponding unsupervised branch-local PC2--PC3 view gives $.872/.875$.
Thus the panel shows that a phase-organized component exists within the Top-4
edit subspace, not that phase explains every Top-4 coordinate.

\begin{figure}[!htbp]
  \centering
  \includegraphics[width=0.88\textwidth]{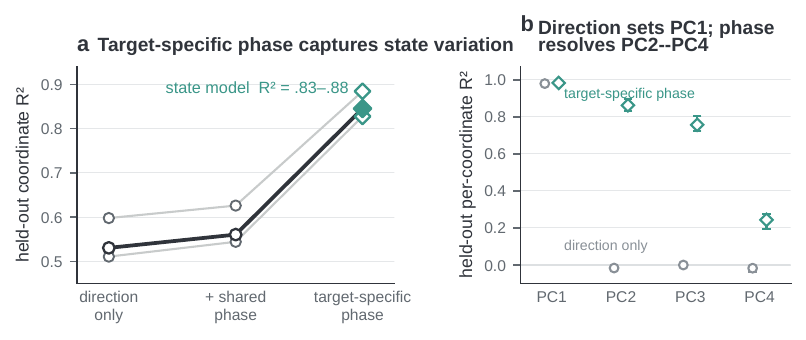}
  \caption{\textbf{Within the tested first-harmonic controller family,
  target-specific boundary phase is needed to predict the Top-4 edit
  coordinates.} \textbf{a}, Thin lines are independently trained checkpoints;
  the thick line is their median.
  Direction-only and a shared additive phase law leave substantial held-out
  error; allowing the first-harmonic coefficients to depend on target direction
  recovers the full state model. \textbf{b}, Direction alone predicts the coarse
  PC1 translation but almost none of PC2--PC4; target-specific phase predicts
  those state-varying coordinates. Markers in \textbf{b} are medians and error
  bars span the minimum and maximum across the three checkpoints.
  All values are fit-only held-out coordinate predictions, not
  decoded recovery rates.}
  \label{fig:boundary-phase-controller-ablation}
\end{figure}

The executable controller in Figure~\ref{fig:causal-route} reads only the
input-history boundary phase
\[
\theta^*=\operatorname{atan2}(-v^*/\omega_{\rm true},x^*)
\]
and requested target direction $s\in\{+1,-1\}$. One pooled fit uses
\[
[1,\cos\theta^*,\sin\theta^*,s,
s\cos\theta^*,s\sin\theta^*]\rightarrow \hat z_{1:4}.
\]
Because $s$ has two values, this is algebraically equivalent to fitting a
separate first-harmonic map $[1,\cos\theta^*,\sin\theta^*]$ for fast- and
slow-target edits. Every model in
Figure~\ref{fig:boundary-phase-controller-ablation} is fit on the same 128
full edits from strict pairs (64 per direction) and tested on 128 disjoint edits.
Direction alone explains the common route offset, but not the state-varying
coordinates; sharing one phase law across both targets also fails. The full
target-specific phase model reaches held-out coordinate $R^2=.828$--$.885$
across three independently trained checkpoints. Adding explicit
$\omega$/amplitude features improves mean held-out $R^2$ by only $.011$, while
raw boundary-state and larger derivative-shaped bases do not improve it.

Repeating the same fit-only phase-aligned-plane analysis on explicit 100K checkpoints
replicates the structure at the localized functional layers B3,
B6, and B4. Qualification is checkpoint-local; held-out fast/slow counts are
$50/45$, $60/53$, and $54/53$ for Runs A--C. Held-out fast/slow phase-plane
$R^2$ is $.868/.817$ for Run A,
$.934/.881$ for Run B, and $.924/.895$ for Run C (315 held-out
paired-difference records total).
For Run B, these values are nearly identical to the 50K result
($.935/.881$), whereas its branch-local PC2--PC3 control at 100K is
$.866/.824$. The contrast shows that apparent loop tightness depends on the
projection: the fit-only phase-aligned plane and unsupervised local PCs answer
different questions within the same frozen route.

\FloatBarrier

\subsection{Boundary-state and after-conflict controllers}
\label{app:state-parameterizations}

The boundary-state controller uses only the input boundary state and requested
rewrite direction to predict $\hat z_{1:4}$ and reconstruct the condition-prefix
edit through the frozen top-four basis. Baseline rollouts define the held-out
strict-failure evaluation cohort but are not inputs to the controller.
A complementary coordinate-difference model first decodes the
natural conflict and estimates its realized frequency and future-start phase,
$(\hat\omega_C,\hat\phi_C)$. Given a requested target state
$(\omega_T,\phi_T)$, it forms
\[
\Psi(\omega,\phi)
=[\omega,\cos\phi,\sin\phi,\omega\cos\phi,\omega\sin\phi],
\qquad
\Delta\Psi_{C\to T}=\Psi(\omega_T,\phi_T)-
\Psi(\hat\omega_C,\hat\phi_C).
\]
The requested physical frequency is mapped into the model's decoded-frequency
convention by a fit-only calibration, and target phase is supplied by a
target-specific first-harmonic law fit on the same training split. On the 128
held-out receivers per run, constructing the edit reads only the natural
conflict rollout---not the held-out aligned video or activation. Held-out
coordinate $R^2$ is $.828/.879/.850$ across Runs A--C, and median decoded
recovery over the six run--direction groups ranges from $.895$ to $1.004$.

This difference form has useful structure. It is zero when target and current
states agree, reverses sign when source and target are exchanged, and composes
additively through intermediate states. The sine--cosine embedding removes the
$2\pi$ phase discontinuity; $\Delta\omega$ supplies the coarse frequency
translation, while the phase and $\omega$-weighted phase terms provide
position-like and velocity-like corrections. A single pooled linear map therefore
supports continuous target specification, although the decoded evaluations in this paper are restricted
to the two matched target families; they do not establish reliable control of
unsupported middle frequencies. The model remains retrospective because
$(\hat\omega_C,\hat\phi_C)$ must be measured from the first generated future;
the target state is specified without another donor video.

\begin{figure}[!htbp]
  \centering
  \includegraphics[width=\textwidth]{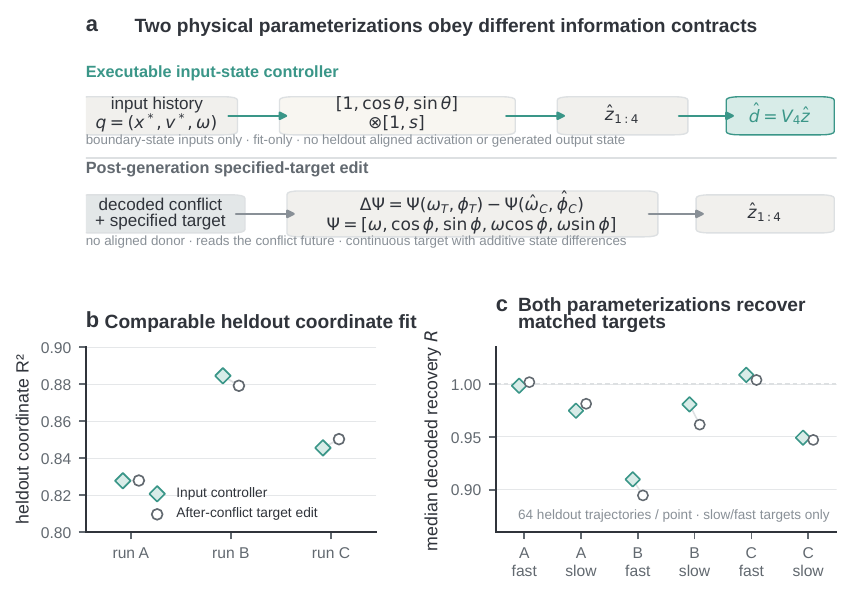}
  \caption{\textbf{Two physical parameterizations describe the route under
  different information contracts.} \textbf{a}, The boundary-state controller
  uses only input boundary phase and target direction to construct its edit. The coordinate-difference
  model measures the decoded conflict state and compares it with a
  target state supplied by fit-only frequency calibration and phase prediction;
  it reads neither the held-out aligned video nor an aligned activation donor.
  \textbf{b}, Both parameterizations predict similar held-out route coordinates.
  \textbf{c}, Injecting either prediction recovers the matched target family,
  but only the boundary-state controller constructs its edit without reading the
  generated conflict future. Decoded interventions are evaluated only for the slow and fast
  matched target families, not unsupported middle frequencies.}
  \label{fig:state-parameterizations}
\end{figure}

\FloatBarrier

\subsection{Full-dimensional mean and receiver-specific edit controls}
\label{app:short50k-fullmean-controls}

\begin{figure}[!htbp]
  \centering
  \includegraphics[width=\textwidth]{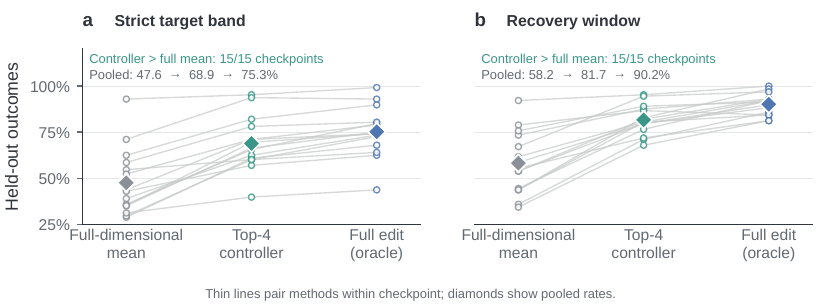}
  \caption{\textbf{State-conditioned Top-4 controllers outperform a
  full-dimensional global mean across 15 checkpoints.} \textbf{a}, Strict
  target-band rates. \textbf{b}, Recovery-window rates. Each checkpoint
  (seeds 3407--3421) contributes 128 disjoint held-out receivers. Thin lines
  pair methods within a checkpoint; small points are checkpoint rates and
  diamonds are pooled rates. The fit-only full mean keeps all activation
  dimensions but is constant across receivers within a target direction. The
  full edit uses each receiver's aligned-minus-conflict difference and is
  therefore an oracle reference, not a transferable controller. All displayed
  intervention records are valid.}
  \label{fig:short50k-fullmean-controls}
\end{figure}

The state-conditioned Top-4 controller is not helped merely by operating in a
larger activation space. We compare it with a fit-only full-dimensional mean
that retains every activation coordinate but is constant across held-out
receivers within each target direction. Across all 15 checkpoints, the
controller has a higher strict target-band rate and a higher recovery-window
rate at every checkpoint. Pooled over the 1,920 held-out receivers, strict
success rises from $47.6\%$ to $68.9\%$, and recovery-window success rises from
$58.2\%$ to $81.7\%$. The receiver-specific full edit reaches $75.3\%$ and
$90.2\%$, respectively, but is an oracle diagnostic because it reads each
held-out aligned activation (Figure~\ref{fig:short50k-fullmean-controls}).

\FloatBarrier

\subsection{Cross-system replications of physically organized writability}
\label{app:cross-system-replications}

\subsubsection{Pendulum}
\label{app:pendulum-replication}

We repeat the behavioral, edit-geometry, decoded-recovery, and depth-localization
analyses in a Pendulum frequency--color task. Table~\ref{tab:pendulum-contract}
summarizes the frozen experimental contract.

\begin{table}[H]
  \centering
  \caption{\textbf{Pendulum replication contract and provenance.}}
  \label{tab:pendulum-contract}
  \begin{tabular}{p{0.27\textwidth}p{0.66\textwidth}}
    \toprule
    Component & Frozen setting \\
    \midrule
    Model & width-1152 DiT, seed 3407, Short/Long 50K \\
    Checkpoint & step 50,000 \\
    Behavior & 64 physical states $\times$ 11 cues in each of four history--band cells; 2,816 decoded futures \\
    Mechanism split & 128 strict receivers; 64 fit and 64 disjoint held-out, balanced 32/32 per target direction \\
    Intervention & after block 12 (zero based), all 20 flow-matching calls, 1,088 condition-prefix tokens; target suffix unchanged \\
    Evaluator & Detection $\ge50\%$, missing runs $\le12$ frames, adjacent/boundary jumps $\le30$ px, and a valid oscillation fit; common across conditions \\
    Same-rank comparison & Paired-difference Top-4 coordinates and controller-predicted Top-4 coordinates; global scale 1.0 selected on fit only \\
    \bottomrule
  \end{tabular}
\end{table}

The frozen oscillation-fit RMSE limit is $.12$; bob size and inferred length
are retained as diagnostics, not exclusion gates.

\begin{figure}[tbp]
  \centering
  \includegraphics[width=\textwidth]{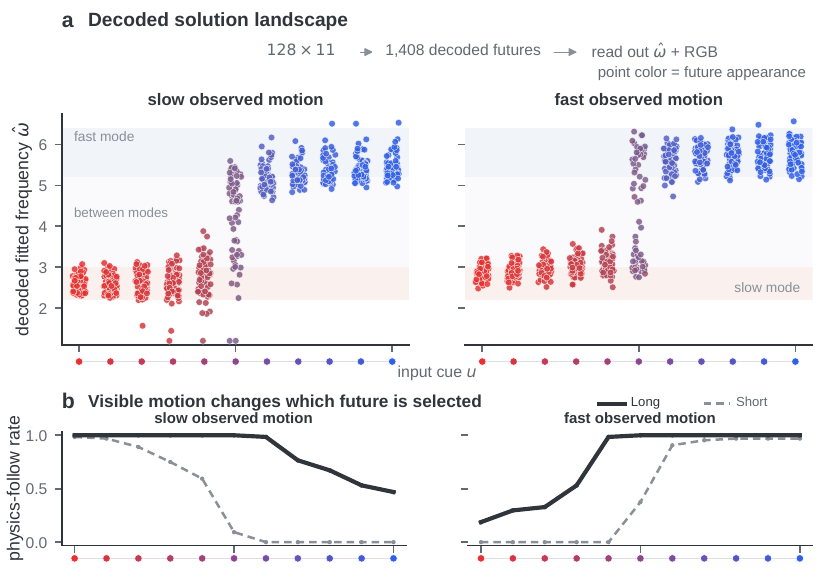}
  \caption{\textbf{Appearance cue and observed motion compete in Pendulum
  decoded rollouts.} \textbf{a}, The Short landscape holds low or high observed
  motion fixed while sweeping 11 cues; point RGB is measured from the generated
  future, and pale bands mark the two training-supported frequency ranges.
  \textbf{b}, The same per-cue physics-follow summary compares Short and Long
  input using the main-paper line-style convention.}
  \label{fig:pendulum-behavior}
\end{figure}

\textbf{Appearance cue and observed motion compete in decoded Pendulum
futures.} All $4\times64\times11=2{,}816$ rollouts yield valid dynamics and
appearance measurements, with physical-state identity as the clustering unit.
Endpoint cues favor their training-associated frequency, while ambiguous cues
give observed motion more control; generated appearance moves with the selected
motion (Figure~\ref{fig:pendulum-behavior}).

\begin{figure}[tbp]
  \centering
  \includegraphics[width=\textwidth]{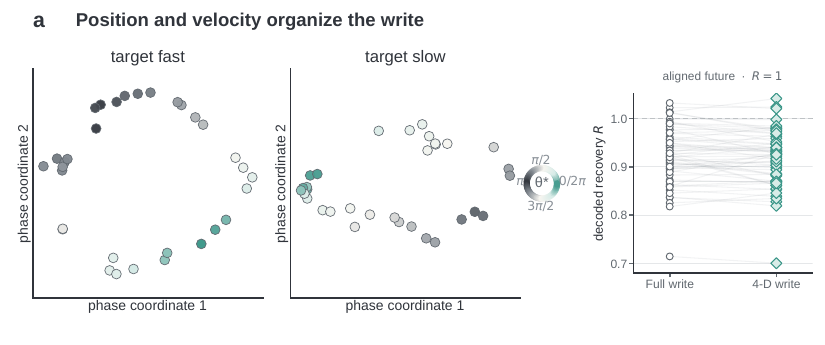}
  \caption{\textbf{Pendulum physical state organizes edit coordinates.}
  The two phase planes show disjoint held-out target-fast and target-slow writes
  in the frozen fit-only coordinate system; point color is boundary phase.
  The paired inset compares full and Top-4 decoded recovery on all 64 held-out
  receivers. Held-out phase-plane $R^2$ is $.985/.975$.}
  \label{fig:pendulum-state-geometry}
\end{figure}

\textbf{Boundary angle and angular velocity organize the edit.} Raw,
uncentered PCA is fit on paired differences from the fit split, and the
top-four coordinates are retained. Within each target direction, a first-harmonic
law of boundary phase---the oscillator-specific encoding of angle and angular
velocity---defines a frozen two-dimensional plane. Disjoint held-out edits lie
on the same structure, with phase-plane $R^2=.975$ for target-low and $.985$ for
target-high ($n=32$ each; Figure~\ref{fig:pendulum-state-geometry}).

\begin{figure}[tbp]
  \centering
  \includegraphics[width=\textwidth]{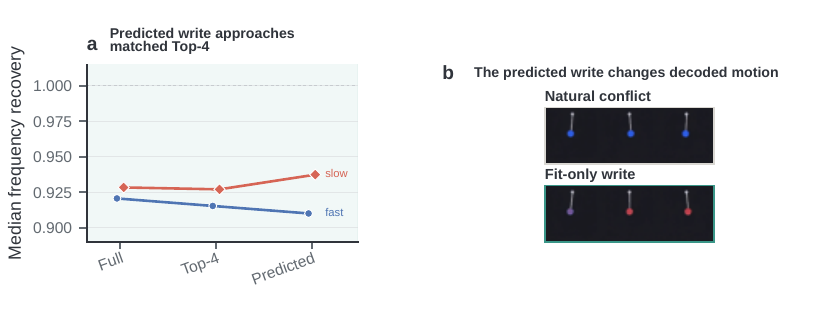}
  \caption{\textbf{A donor-free top-four Pendulum controller approaches the
  Top-4 reference.} \textbf{a}, Fast- and slow-target medians among
  evaluator-valid receivers compare Full, Top-4, and predicted
  writes in the same coordinate space. \textbf{b}, Real decoded strips show the
  same held-out receiver before and after the fit-only write. All-cohort validity
  and near-full counts are reported in the accompanying text.}
  \label{fig:pendulum-decoded-recovery}
\end{figure}

\textbf{A donor-free controller reaches the same-rank reference.} Across 64
held-out receivers, the natural-conflict, full-edit, Top-4, and predicted-edit
conditions are evaluator-valid for all $64$ receivers; their all-cohort
near-full counts are $0$, $63$, $63$, and $63$. Median recovery is $0$, $.925$,
$.926$, and $.917$, respectively. The fit-only controller therefore approaches
its Top-4 reference while using fit-learned parameters and held-out boundary
state. It produces 58 physics-following futures and six compromises. The decoded
frame strips hold receiver, noise, schedule, and future window fixed
(Figure~\ref{fig:pendulum-decoded-recovery}).

\begin{figure}[tbp]
  \centering
  \includegraphics[width=\textwidth]{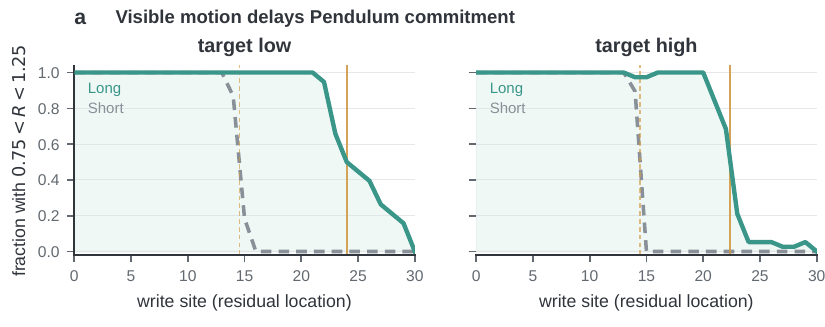}
  \caption{\textbf{Visible motion delays Pendulum commitment in both directions.}
  Target-low and target-high directions are shown separately with Short and
  Long common-strict rewrite-probability curves overlaid. Thin amber lines mark
  their separate $L_{50}$ closure sites.}
  \label{fig:pendulum-writability}
\end{figure}

\textbf{Writability again closes over a localized depth range.} On the
Short/Long common-strict cohort ($n=38$ paired receivers per direction), Long
increases integrated writability $D^\omega$ by $10.53$ sites for target-low
and $8.05$ sites for target-high. The midpoint $L_{50}$ moves later by $9.46$
and $7.95$ sites, respectively. Additional motion history therefore delays
commitment in both directions (Figure~\ref{fig:pendulum-writability}).

\FloatBarrier

\subsubsection{Free Fall}
\label{app:freefall-replication}

Free Fall tests the same claims in a non-oscillatory system. A mass is released
from rest at a fixed boundary state, with red paired with the low-gravity range
$g\in[.003,.025]$ and blue with the high-gravity range
$g\in[.045,.067]$. The model has 30 DiT blocks and is evaluated at 100K with
32 visible conditioning frames. Table~\ref{tab:freefall-contract} summarizes
the frozen contract.

\begin{table}[H]
  \centering
  \caption{\textbf{Free-fall replication contract and provenance.}}
  \label{tab:freefall-contract}
  \begin{tabular}{p{0.27\textwidth}p{0.66\textwidth}}
    \toprule
    Component & Frozen setting \\
    \midrule
    Model & width-1536 DiT, 30 blocks, 12 heads, seed 3407, 100K \\
    Behavior & 64 Short-regime trajectories $\times$ 11 cues; 704 decoded futures displayed. The full 1,408-row Short/Long diagnostic remains in source data \\
    Mechanism bank & 128 gravity-error pairs, balanced across low/high gravity: aligned $|\hat g-g|<.002$, conflict $|\hat g-g|\ge.002$ \\
    Intervention & after block 1, all 20 flow-matching calls; observed-frame tokens edited and future-frame tokens unchanged at insertion \\
    PCA and controller & Shared uncentered PCA fit on 64 pairs only; direction-specific $[1,g_{\mathrm{target}}]$ laws fit on 32 pairs each. The remaining 64 pairs (32 per direction) are held out, with disjoint base seeds \\
    Evaluator & Gravity-fit success: $|\hat g-g|<.002$ \\
    \bottomrule
  \end{tabular}
\end{table}

Here failure means inaccurate gravity, not necessarily a switch to the opposite
gravity interval. Success and normalized recovery
$R=(\hat g_{\mathrm{edit}}-\hat g_C)/(\hat g_A-\hat g_C)$ use the same fitted
gravities; $A$ and $C$ denote the aligned and conflict baselines.

\begin{figure}[H]
  \centering
  \includegraphics[width=\textwidth]{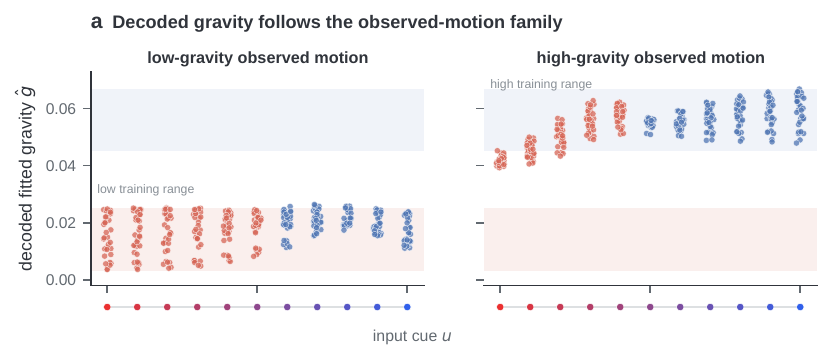}
  \caption{\textbf{Decoded Free Fall gravity follows the observed-motion
  family.} Holding low- or high-gravity observed motion fixed in the 32-frame
  regime, we sweep 11 appearance cues and fit $\hat g$ from all 704 decoded
  futures. Point color is measured future appearance; pale bands show the two
  training-supported gravity ranges.}
  \label{fig:freefall-behavior}
\end{figure}

\textbf{Decoded gravity remains in the observed-motion family across appearance
cues.} We display the 704 decoded futures from the 32-frame Short regime used by
the mechanism experiments. In both physical ranges, hue shifts fitted gravity
within the history-consistent family without switching to the other range
(Figure~\ref{fig:freefall-behavior}). Point color shows decoded appearance.

\begin{figure}[H]
  \centering
  \includegraphics[width=\textwidth]{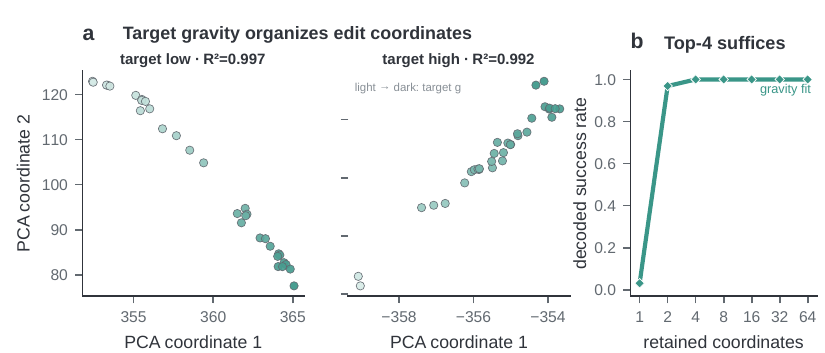}
  \caption{\textbf{Target gravity organizes compact Free Fall edit
  coordinates.} \textbf{a}, Held-out paired differences (32 per direction) are shown in
  the first two coordinates of a PCA basis fit on the separate 64-pair fit set;
  point shade is target gravity, and the displayed
  $R^2$ evaluates the direction-specific $[1,g_{\mathrm{target}}]$ law.
  \textbf{b}, Paired-difference projection saturates rapidly with retained PCA
  rank; gravity-fit success rates use the same 64 held-out pairs at every rank.
  Basis and rank-dependent global scales use only the fit set.}
  \label{fig:freefall-geometry}
\end{figure}

\textbf{Specified gravity organizes a compact edit family.} A shared uncentered
PCA basis is fit on 64 paired differences, excluding all 64 held-out pairs.
Within each target direction, a $[1,g_{\mathrm{target}}]$ law is fit on 32 pairs
and evaluated on the other 32. Rank 2 is the smallest rank retaining at least
99\% of fit-set residual energy (99.545\%). At each retained rank, a single
global scale is the square root of total divided by retained fit-set energy;
at rank 2 it is $1.00228$, shared by oracle projection and controller, with no
per-receiver norm matching. Held-out joint coordinate $R^2$ is $.997/.992$
for low/high targets. Four-coordinate oracle projection achieves gravity-fit
success on all 64 held-out pairs
(Figure~\ref{fig:freefall-geometry}).

\begin{figure}[tbp]
  \centering
  \includegraphics[width=\textwidth]{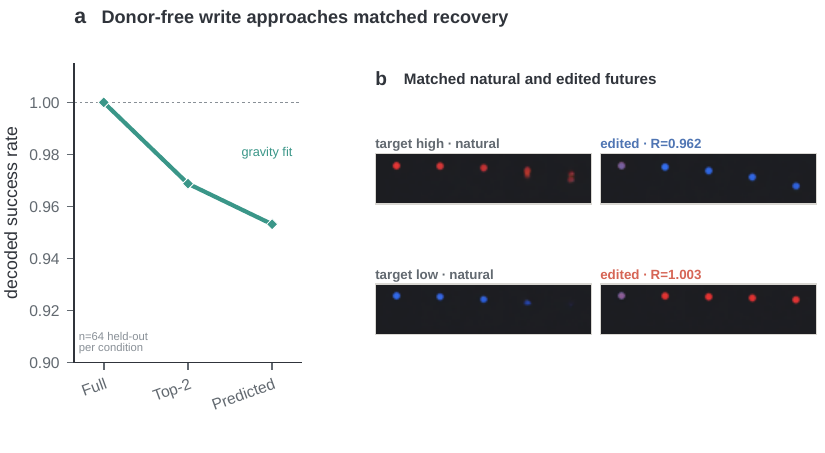}
  \caption{\textbf{A donor-free free-fall controller approaches matched
  decoded recovery.} \textbf{a}, Gravity-fit success is reported for the
  full-edit reference, Top-2 oracle projection, and predicted directional
  controller on the same 64 held-out receivers (64, 62, and 61 successes,
  respectively). \textbf{b}, Real decoded
  keyframes show matched natural and controller-edited held-out examples with the
  same receiver, generation seed, flow-matching schedule, and future window.}
  \label{fig:freefall-decoded-recovery}
\end{figure}

\textbf{The donor-free controller changes held-out decoded futures.} Frozen
direction-specific $[1,g_{\mathrm{target}}]$ controllers map specified gravity
into the same two-coordinate space without reading a held-out activation donor.
All 64 held-out outputs are evaluator-valid; median normalized recovery is
$.967$; gravity-fit success is $61/64=.953$, with mean absolute
gravity error $.00086$. The edit also moves appearance toward the
training-associated target color in all 64 outputs, so this is a physically
parameterized edit to the coupled target future, not a color-disentangled
gravity subspace. Figure~\ref{fig:freefall-decoded-recovery} includes two real
matched natural/controller-edited held-out comparisons, selecting the receiver
closest to the direction-specific median recovery in each target range.

\begin{figure}[tbp]
  \centering
  \includegraphics[width=\textwidth]{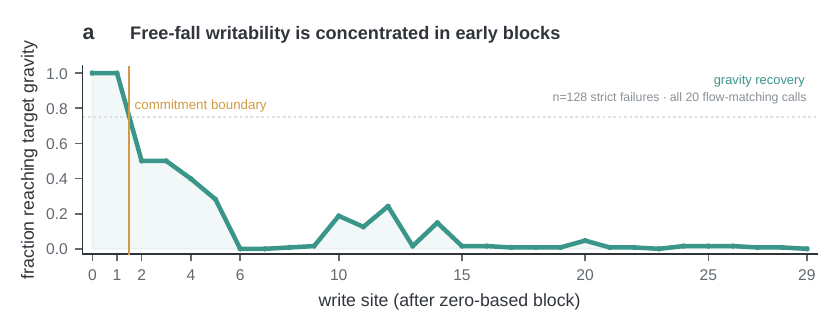}
  \caption{\textbf{Free Fall reproduces localized causal writability.}
  Gravity-fit success is shown at every intervention block on the same 128
  gravity-error pairs. The amber line marks the $.75$ success-fraction boundary
  between B1 and B2; every edit is applied at all 20 flow-matching calls.}
  \label{fig:freefall-writability}
\end{figure}

\textbf{Direct writability is concentrated in early blocks.} The same full
edit reaches gravity-fit success of $1.0$ after blocks 0 and 1, then the
gravity-fit success fraction falls below $.75$ after block 1
(Figure~\ref{fig:freefall-writability}). The full 30-block profile is displayed.

\FloatBarrier

\subsection{Persistent-error composition control}
\label{app:persistent-rigidity}

\begin{figure}[!htbp]
  \centering
  \includegraphics[width=\textwidth]{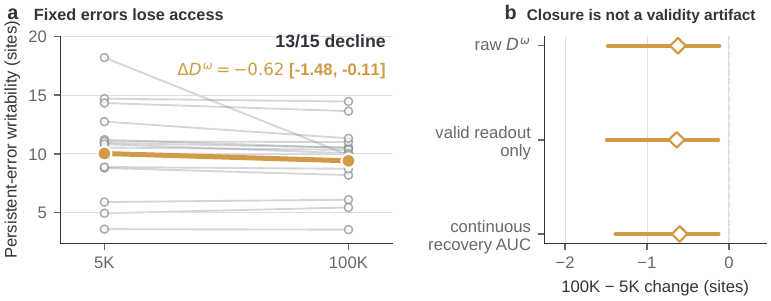}
  \caption{\textbf{The same unresolved shortcut errors become less causally
  writable during training.} \textbf{a}, Each gray line is one seed cohort
  evaluated on fixed persistent-error trajectory identities; amber is the
  trajectory-weighted mean. Thirteen of fifteen cohorts decline from 5K to
  100K, with mean paired change $\Delta D^\omega=-0.62$ sites
  (seed-clustered 95\% CI $[-1.48,-0.11]$). \textbf{b}, The same 886 trajectories
  lose access under raw, valid-only, and continuous-recovery scoring, using the
  same trajectory weighting and seed-then-trajectory bootstrap.}
  \label{fig:persistent-rigidity}
\end{figure}

Figure~\ref{fig:persistent-rigidity} removes a possible sample-composition
explanation for the training-time contraction in causal writability.  We
freeze the trajectory identities that remain strict shortcut failures at all
six evaluated checkpoints, then compare those same trajectories at 5K and
100K.  Thus a later checkpoint cannot appear less writable merely because its
local strict bank contains a different set of failures.

\FloatBarrier

\subsection{Writability robustness and history dependence}
\label{app:writability-robustness}

We test whether the writability results depend on the near-full recovery
threshold, measured output/state covariates, or invalid frequency readouts. We
then report the operational Short/Long boundaries separately for each of the
three training seeds.

\begin{figure}[!htbp]
  \centering
  \includegraphics[width=\textwidth]{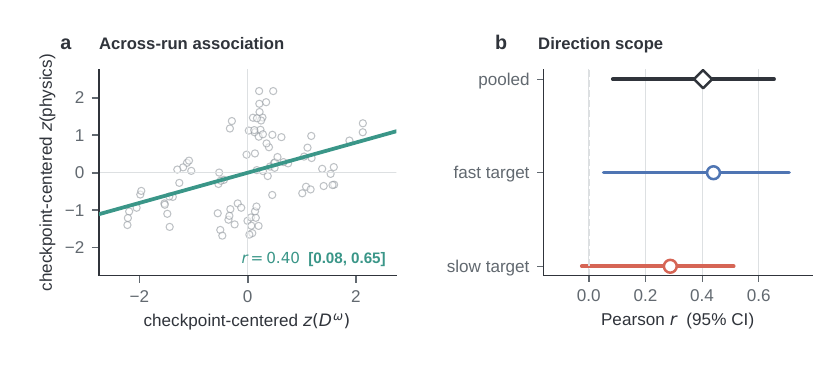}
  \caption{\textbf{Across-run writability is associated with physics behavior
  at the same training steps.} \textbf{a}, After checkpoint centering, runs with
  larger $D^\omega$ tend to be more physics-following. \textbf{b}, The
  association is clearer for fast-target quantities than for the slow
  target; the pooled result summarizes a direction-asymmetric tendency.
  Complete six-checkpoint seed trajectories are the
  bootstrap unit.}
  \label{fig:writability-cross-solution}
\end{figure}

The checkpoint-centered cross-solution association is direction-asymmetric:
fast-target writability has $r=.44$ (95\% CI $[.05,.71]$), whereas the
slow-target estimate is $r=.29$ (95\% CI $[-.03,.51]$). The pooled
$r=.40$ association in Figure~\ref{fig:writability-cross-solution}a summarizes an overall
tendency, with a stronger estimate in the fast branch.

\begin{figure}[!htbp]
  \centering
  \includegraphics[width=\textwidth]{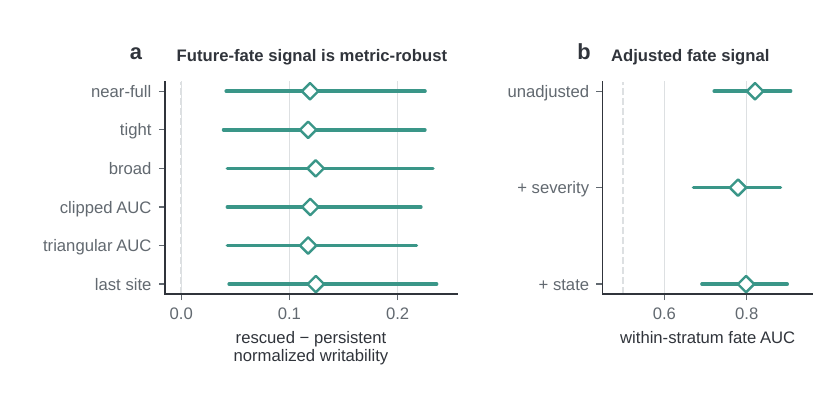}
  \caption{\textbf{Early writability and training-time closure survive
  alternative definitions and measured controls.} \textbf{a}, The early gap
  between future-rescued and persistent failures remains under six definitions
  of writability (157 rescued, 953 persistent). \textbf{b}, The same early
  separation remains after controlling natural error severity and boundary
  state. Diamonds show pooled trajectory mean differences in \textbf{a} and
  direction-balanced AUC in \textbf{b}; horizontal lines are
  seed-clustered trajectory bootstrap 95\% intervals.}
  \label{fig:writability-robustness}
\end{figure}

\begin{figure}[!htbp]
  \centering
  \includegraphics[width=\textwidth]{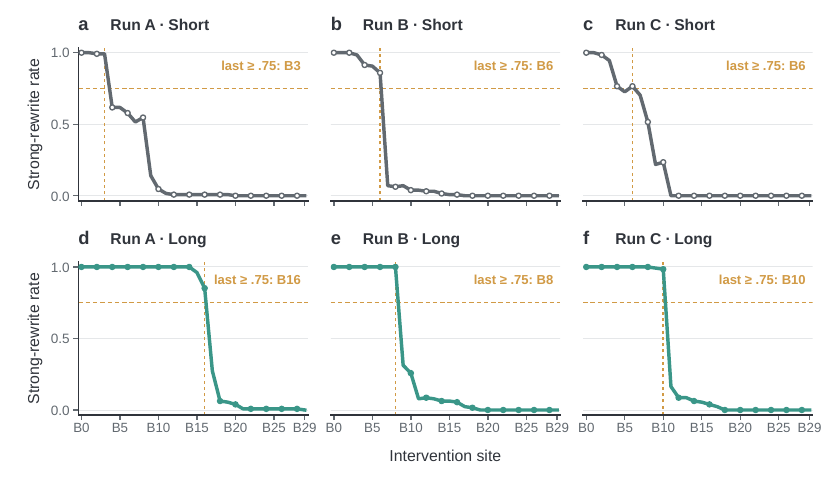}
  \caption{\textbf{Long-history fast-target writes remain effective deeper in
  all three Spring solutions.} \textbf{a--c}, Short-history layer scans for
  Runs A--C. \textbf{d--f}, The corresponding Long-history scans. Every panel
  shows the complete checkpoint-specific 128-pair strict bank at 50K; each
  point is the fraction with a valid decoded output and
  $.75<R^\omega<1.25$. The amber horizontal line marks the $.75$ operational
  threshold, and the vertical line marks the last after-block site meeting it.
  Short and Long use the same architecture, checkpoint step, intervention, and
  readout protocol, but their strict banks are not trajectory-paired.}
  \label{fig:spring-short-long-layer-scans}
\end{figure}

Figure~\ref{fig:spring-short-long-layer-scans} expands the Short/Long boundary
summary into all six complete fast-target layer profiles. We restrict this
comparison to the fast target because no complete three-run Long slow-target
strict cohort is available: Runs B and C each qualified zero of 4,096 evaluated
candidates, and no corresponding Run-A qualification result was located in the
audited Large-Long result root.

\begin{table}[!htbp]
  \centering
  \caption{\textbf{Long checkpoints have later operational fast-target write
  boundaries for all three training seeds at 50K.} Boundaries are the last
  after-block sites whose strong-rewrite rate is at least $.75$. Short and Long
  use the same architecture, step, and intervention protocol, but each
  checkpoint is evaluated on its own 128-pair strict bank; trajectory identities
  are not paired across history regimes.}
  \label{tab:short-long-boundaries}
  \begin{tabular}{lrrr}
    \toprule
    Run & Short boundary & Long boundary & Long $-$ Short \\
    \midrule
    A & B3 & B16 & $+13$ \\
    B & B6 & B8 & $+2$ \\
    C & B6 & B10 & $+4$ \\
    \bottomrule
  \end{tabular}
\end{table}

\FloatBarrier

\subsection{Early error fate and downstream target write}
\label{app:fate-target-write}

\begin{figure}[!htbp]
  \centering
  \includegraphics[width=\textwidth]{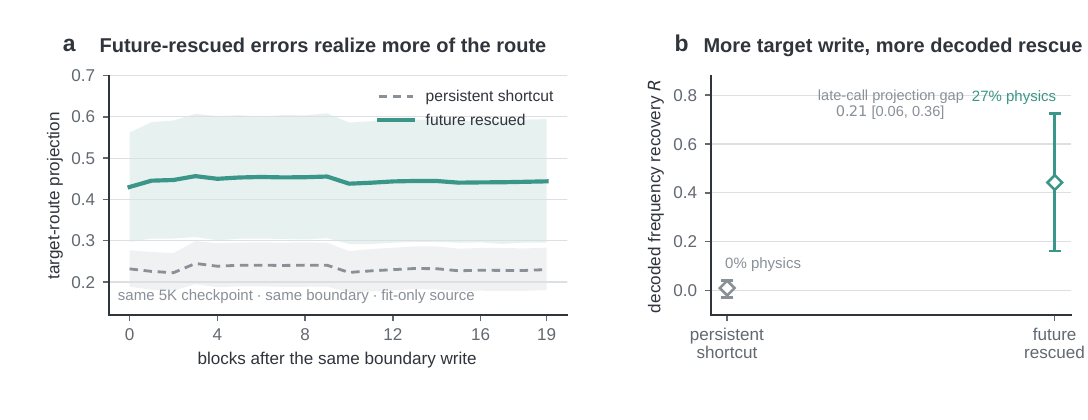}
  \caption{\textbf{Early failures later rescued by training already realize
  more of the condition-to-target route.} At the same 5K boundary, trajectories
  later rescued by training already show a larger route-aligned target-state
  response than persistent failures (\textbf{a}). A K-only intervention also
  changes decoded frequency only in the future-rescued group (\textbf{b}; 11
  rescued and 12 persistent fast-target failures). The source controller is fit
  on separate trajectories. Curves and markers show trajectory means; shaded
  bands and error bars are trajectory-bootstrap 95\% intervals.}
  \label{fig:fate-target-write}
\end{figure}

Figure~\ref{fig:fate-target-write} asks whether trajectories with different
future training outcomes already differ in downstream route realization at the
same early checkpoint and boundary.

\FloatBarrier

\subsection{Cross-run transfer controls and target-conditioned outcomes}
\label{app:cross-run-transfer-controls}

\begin{figure}[!htbp]
  \centering
  \includegraphics[width=\textwidth]{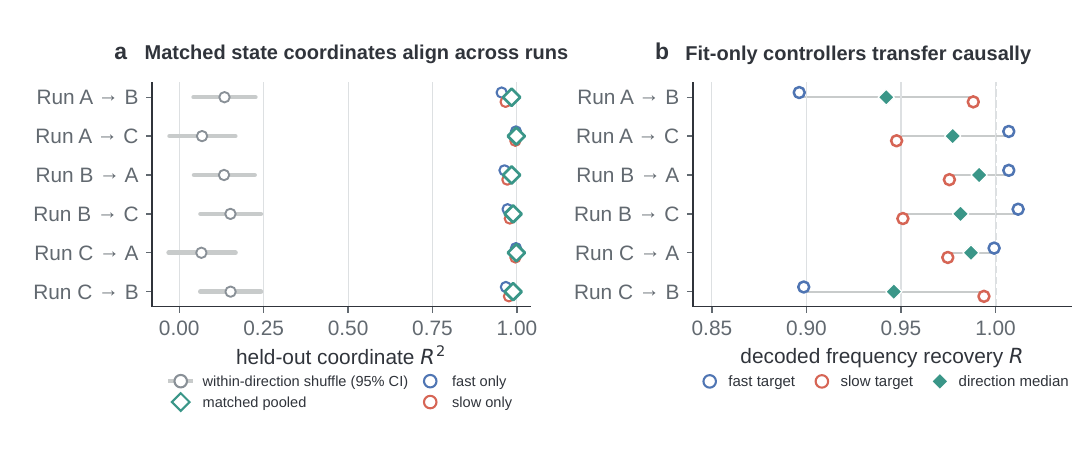}
  \caption{\textbf{Internal geometry transfers across independently
  trained models.} \textbf{a}, A fit-only scale and four-dimensional orthogonal
  map aligns held-out matched coordinates in all six directed run pairs;
  within-direction trajectory shuffles do not. \textbf{b}, Source-controller
  predictions transferred through the frozen maps recover decoded frequency in
  both rewrite directions. The shuffle in \textbf{a} is a coordinate null; the
  decoded claim in \textbf{b} comes from separate held-out interventions.}
  \label{fig:cross-run-transfer-controls}
\end{figure}

The scale-plus-orthogonal maps in Figure~\ref{fig:cross-run-transfer-controls}
are fit on 128
matched differences and evaluated on 128 disjoint physical identities for every
directed run pair. Matched within-direction coordinate $R^2$ ranges from $.955$
to $.997$, whereas shuffling trajectory identity within direction gives means
of $.066$--$.152$; none of 20,000 shuffles reaches the matched value
($p<0.00005$ for every map). This shuffle tests coordinate geometry,
not decoded intervention efficacy. The separate intervention test passes held-out
boundary state through the source controller, frozen cross-run map, and target
basis. All six directed maps redirect decoded frequency with pooled
direction-median recovery $R=.942$--$.991$.

Absolute physics-follow rates remain bounded by the target model's own full-edit
ceiling; normalized recovery is therefore the primary cross-run transfer
endpoint.

\FloatBarrier

\subsection{Matched-coordinate compatibility across training checkpoints}
\label{app:cross-step-exact100}

We next ask whether a route coordinate selected at one training checkpoint can
be expressed at another checkpoint in the same run. The assay uses exactly 5K,
10K, 20K, 50K, 80K, and 100K. Every checkpoint independently qualifies strict
failures from the same frozen 256-trajectory library. Each source--target map is
fit on the qualified intersection inside one permanent fit split and evaluated
on the corresponding intersection inside the globally disjoint held-out split;
a physical identity never changes split.

\begin{figure}[!htbp]
  \centering
  \includegraphics[width=\textwidth]{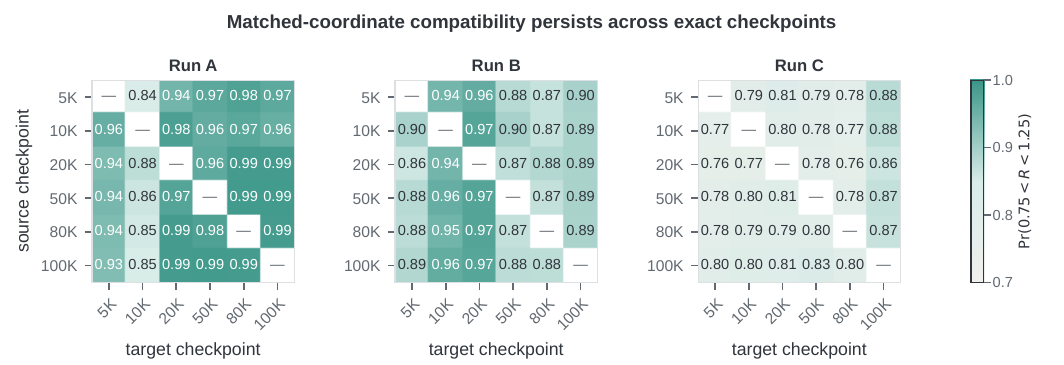}
  \caption{\textbf{Matched route coordinates remain compatible across
  exact training checkpoints.} Cells report the held-out fraction satisfying
  $.75<R<1.25$ after a Top-4 coordinate selected at the source
  checkpoint is mapped by a fit-only
  scaled-orthogonal transform and injected at the target checkpoint's own
  rewrite site. All 90 directed off-diagonal maps are shown; each cell contains
  88--123 held-out physical trajectories, with Wilson 95\% intervals in the
  source data. Blank diagonals mean self-transfer was not evaluated. The assay
  tests within-run temporal compatibility, not the predicted controller
  transfer in Figure~\ref{fig:cross-run-transfer-controls}b.}
  \label{fig:cross-step-exact100}
\end{figure}

Compatibility is high but solution-dependent: the median cell fraction is
$.966$, $.893$, and $.796$ for Runs A--C, respectively. This does not contradict
contracting writability. Figure~\ref{fig:cross-step-exact100} asks whether a
matched selected coordinate can be transported between each checkpoint's own
rewrite site on a qualified intersection; $D^\omega$ asks how long a local
condition-side edit remains directly effective across depth.

\FloatBarrier

\subsection{Shared mediation and implementation multiplicity}
\label{app:implementation-multiplicity}

Having established shared route geometry, we compare separately calibrated
head- and component-level interventions. The goal is to identify how the edit
is mediated within each run, not to rank components at a
common gain.

\begin{figure}[!htbp]
  \centering
  \includegraphics[width=0.92\textwidth]{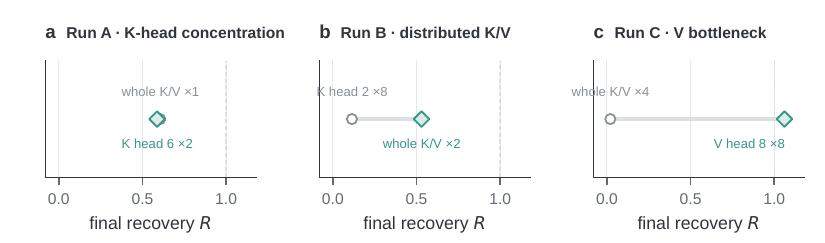}
  \caption{\textbf{Shared route semantics admit distinct downstream
  bottlenecks under the tested interventions.} In Run A, one K head closely
  reproduces the calibrated whole-K/V response. In Run B, the tested K-head
  intervention at gain 8 remains below whole-K/V at gain 2. In Run C, the tested
  whole-K/V gains 1, 2, and 4 do not cross the margin, whereas the selected
  V-head intervention does. Markers show mean final recovery over 16 held-out
  trajectories per panel. The panels establish implementation multiplicity.
  Independently calibrated doses do not support equal-dose component rankings.}
  \label{fig:implementation-multiplicity}
\end{figure}

\begin{figure}[H]
  \centering
  \includegraphics[width=\textwidth]{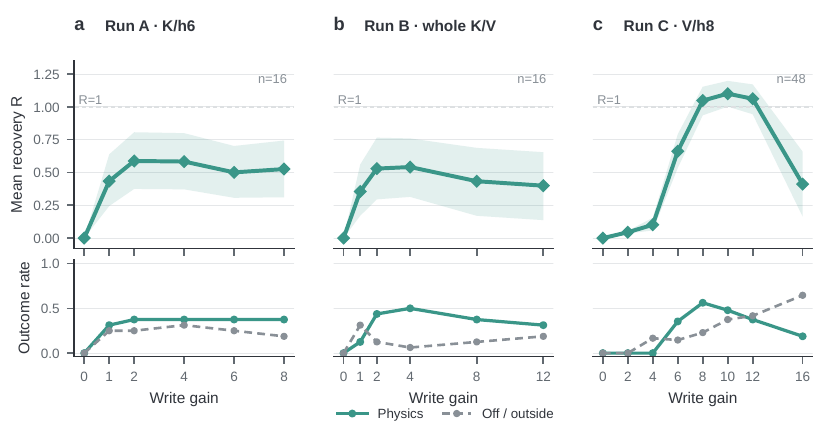}
  \caption{\textbf{Distinct K/V implementations have distinct causal dose
  responses.} \textbf{a}, Run A writes only K head 6 at gains 1, 2, 4, 6, and
  8 on 16 common selection-clean held-out fast-target failures. \textbf{b}, Run
  B writes whole condition K/V at gains 1, 2, 4, 8, and 12 on the same 16
  physical identities. \textbf{c}, Run C writes only V head 8 at gains 2, 4,
  6, 8, 10, 12, and 16 on 48 selection-clean held-out failures. Upper panels
  show mean frequency recovery with trajectory-bootstrap 95\% intervals; lower
  panels show physics-follow and off/outside-family fractions. Gain zero is the
  unedited-conflict reference, for which $R=0$ by definition. Every executed
  arm is evaluator-valid, uses all 20 flow-matching calls, and does not read a
  held-out aligned activation. Operators and gain scales differ across panels,
  so the figure establishes solution-specific causal margins rather than an
  equal-dose component ranking.}
  \label{fig:short-attention-gain-sweeps}
\end{figure}

The compact comparison above uses independently calibrated doses. To expose the
underlying causal margins, Figure~\ref{fig:short-attention-gain-sweeps} reports
the complete tested gain grids for the intervention that is informative in each
run.

\begin{table}[!htbp]
  \centering
  \caption{\textbf{Component restoration localizes a common attention-mediated
  target write.} The first four rows restore natural-conflict condition
  components inside the successful full edit; effects are relative to the
  intact edit over six run--direction groups with 16 paired receivers each. The
  final rows compare post-attention target-state and MLP-output interventions.
  Intervals are hierarchical-bootstrap 95\% CIs over group medians.}
  \label{tab:qkv-attention-mlp}
  \setlength{\tabcolsep}{5pt}
  \begin{tabular}{@{}p{0.43\textwidth}lrr@{}}
    \toprule
    Paired contrast & Unit & $\Delta R$ & 95\% CI \\
    \midrule
    Restore conflict Q $-$ full edit & $6\times16$ & $+.0004$ & $[-.0010,+.0047]$ \\
    Restore conflict K $-$ full edit & $6\times16$ & $-.917$ & $[-.967,-.028]$ \\
    Restore conflict V $-$ full edit & $6\times16$ & $-.003$ & $[-.456,+.005]$ \\
    Restore conflict K+V $-$ full edit & $6\times16$ & $-.931$ & $[-.985,-.908]$ \\
    Edited post-attention target $-$ edited MLP output & 23 receivers / 6 groups & $+.960$ & $[+.253,+1.024]$ \\
    Restore conflict post-attention target $-$ restore conflict MLP output & 24 receivers / 6 groups & $-.979$ & $[-1.042,-.938]$ \\
    \bottomrule
  \end{tabular}
\end{table}

\FloatBarrier

\subsection{Specificity and same-cohort full-edit references}
\label{app:single-head-specificity}

For the Run-C V/h8 bottleneck, we test whether decoded rescue specifically tracks
realization of the target route. We also compare the
same 48 failures under the B9 head intervention and three B4 residual-edit
references while keeping their distinct operators explicit.

\begin{figure}[!htbp]
  \centering
  \includegraphics[width=\textwidth]{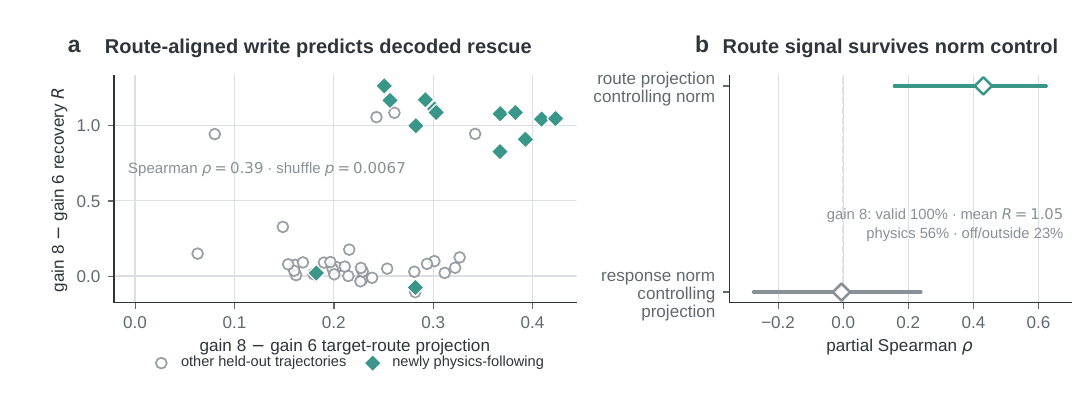}
  \caption{\textbf{Decoded rescue follows the route-aligned target write, not
  generic activation growth.} \textbf{a}, On the same 48 trajectories, the
  gain-6 to gain-8 increase in target-route projection tracks frequency recovery
  and identifies newly physics-following rollouts. \textbf{b}, Route projection
  remains associated with recovery after controlling response norm; the
  estimated partial association of response norm after controlling route
  projection is near zero ($\rho=-.006$, 95\% CI $[-.276,.238]$).}
  \label{fig:single-head-specificity}
\end{figure}

\begin{table}[!htbp]
  \centering
  \caption{\textbf{Single-head rescue is comparable with full-edit references
  on the same 48 failures.} All interventions use the same held-out Run-C
  fast-target cohort, but the first three act on the B4 condition residual,
  whereas the single-head intervention acts on attention V at B9. We report
  physics-follow rates descriptively because these interventions act at
  different sites and through different operators. All four arms are
  evaluator-valid; off-family denotes valid tracks outside the supported
  frequency categories.}
  \label{tab:single-head-oracle-ceiling}
  \setlength{\tabcolsep}{1.8pt}
  \begin{tabular}{@{}llrrrrrr@{}}
    \toprule
    Intervention & Site / operator & \shortstack{Mean\\$R$} & \shortstack{Median\\$R$} & Physics & Shortcut & Compromise & \shortstack{Off-\\family} \\
    \midrule
    Full paired difference & B4 condition residual & $.992$ & $1.002$ & $50.0\%$ & $0.0\%$ & $39.6\%$ & $10.4\%$ \\
    Top-four projection & B4 condition residual & $.992$ & $1.003$ & $41.7\%$ & $0.0\%$ & $41.7\%$ & $16.7\%$ \\
    Fit-only state edit & B4 condition residual & $.972$ & $.994$ & $43.8\%$ & $0.0\%$ & $41.7\%$ & $14.6\%$ \\
    Single V/h8, gain 8 & B9 attention V head & $1.048$ & $1.087$ & $56.3\%$ & $0.0\%$ & $20.8\%$ & $22.9\%$ \\
    \bottomrule
  \end{tabular}
\end{table}

\FloatBarrier

\subsection{FM-time allocation and decoded single-head rescue}
\label{app:fm-time-rescue}

Finally, we distribute the same gain-8 V/h8 write over different
flow-matching-call windows and inspect both the immediate target-route response
and the decoded continuation.

\begin{figure}[!htbp]
  \centering
  \includegraphics[width=0.92\textwidth]{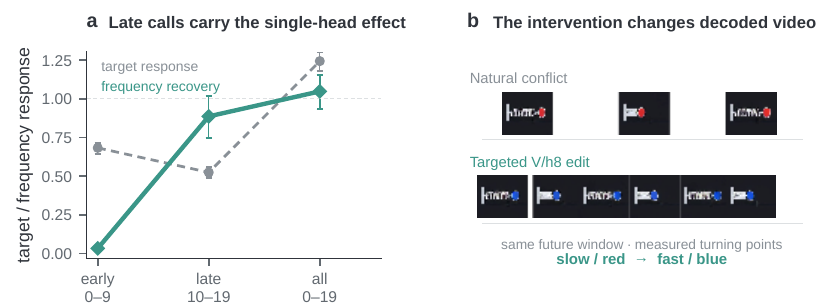}
  \caption{\textbf{The localized V-head write changes decoded video when
  applied over late flow-matching calls.} \textbf{a}, On the same 48
  selection-clean Run-C fast-target failures, an early-call write is internally
  visible but behaviorally ineffective, while late or all-call allocation
  rescues frequency. Markers show trajectory means with trajectory-bootstrap
  95\% intervals. A separate same-identity fast-target cohort ($n=16$) shows
  that neither five-call half of the late window rescues alone. \textbf{b}, A
  representative exact replay changes the decoded joint continuation from
  red/slow to blue/fast; it is the robust-$L_1$ medoid selected from 28 eligible
  successful replays, and frames mark measured turning points in the same future
  window.}
  \label{fig:fm-time-decoded-rescue}
\end{figure}

\FloatBarrier

\FloatBarrier

\end{document}